\documentclass[11pt,openright,twoside]{book}

\usepackage[utf8]{inputenc}
\usepackage[T1]{fontenc}
\usepackage{lmodern}
\usepackage[a4paper,margin=1in]{geometry}
\usepackage{graphicx}
\usepackage{xcolor}
\usepackage{tikz}
\usepackage{microtype}
\usepackage{setspace}
\usepackage{titlesec}
\usepackage{fancyhdr}
\usepackage{emptypage}
\usepackage{epigraph}
\usepackage[round,authoryear]{natbib}
\usepackage{hyperref}

\usepackage{amsmath}
\usepackage{amssymb}
\usepackage{amsthm}

\definecolor{coverdark}{HTML}{0B1E3F}
\definecolor{coveraccent}{HTML}{C9A227}
\definecolor{coverlight}{HTML}{E8ECF2}

\hypersetup{
    colorlinks=true,
    linkcolor=coverdark,
    urlcolor=coverdark,
    citecolor=coverdark,
    pdftitle={Artificial Adaptive Intelligence},
    pdfauthor={Boris Kriuk}
}

\newcommand{\booktitle}{Artificial Adaptive Intelligence}
\newcommand{\booksubtitle}{The Missing Stage Between Narrow and General Intelligence}
\newcommand{\bookauthor}{Boris Kriuk}

\begin{document}

\begin{titlepage}
\thispagestyle{empty}
\begin{tikzpicture}[remember picture, overlay]
    \fill[coverdark] (current page.south west) rectangle (current page.north east);
    \foreach \i in {0,1,...,12}{
        \draw[coveraccent, line width=0.3pt, opacity=0.25]
            ([xshift=-2cm,yshift=\i*1.2cm]current page.south west)
            -- ([xshift=2cm,yshift=\i*1.2cm+6cm]current page.south west);
    }
    \fill[coveraccent] ([yshift=-3cm]current page.north west)
        rectangle ([yshift=-3.08cm]current page.north east);
    \fill[coveraccent] ([yshift=3cm]current page.south west)
        rectangle ([yshift=3.08cm]current page.south east);
    \node[coveraccent, font=\small\scshape, anchor=north]
        at ([yshift=-1.8cm]current page.north) {A Monograph};
    \node[coverlight, font=\fontsize{42}{48}\selectfont\bfseries,
          align=center, text width=15cm]
        at ([yshift=2cm]current page.center) {\booktitle};
    \draw[coveraccent, line width=1.2pt]
        ([xshift=-4cm, yshift=-0.2cm]current page.center)
        -- ([xshift=4cm, yshift=-0.2cm]current page.center);
    \node[coverlight, font=\Large\itshape,
          align=center, text width=14cm]
        at ([yshift=-1.5cm]current page.center) {\booksubtitle};
    \node[coveraccent, font=\Large\scshape, anchor=south]
        at ([yshift=4cm]current page.south) {\bookauthor};
    \node[coverlight, font=\small, anchor=south]
        at ([yshift=3.4cm]current page.south) {2026};
\end{tikzpicture}
\end{titlepage}

\thispagestyle{empty}
\vspace*{6cm}
\begin{center}
    {\Huge\bfseries \booktitle}\\[0.5cm]
    {\large\itshape \booksubtitle}
\end{center}
\cleardoublepage

\thispagestyle{empty}
\vspace*{3cm}
\begin{center}
    {\Huge\bfseries \booktitle}\\[0.8cm]
    \rule{0.6\textwidth}{0.8pt}\\[0.6cm]
    {\Large\itshape \booksubtitle}\\[3cm]
    {\Large \bookauthor}\\[6cm]
    {\small\scshape An Independent Monograph}\\[0.2cm]
    {\small 2026}
\end{center}
\cleardoublepage

\thispagestyle{empty}
\vspace*{\fill}
\noindent
\textit{Artificial Adaptive Intelligence: The Missing Stage Between Narrow and General Intelligence}\\[0.3cm]
Copyright \copyright\ 2026 by Boris Kriuk.\\[0.3cm]
All rights reserved. No part of this publication may be reproduced, stored in a retrieval system, or transmitted, in any form or by any means, without the prior written permission of the author, except in the case of brief quotations embodied in critical articles and reviews.\\[0.6cm]

\noindent First edition, 2026.\\[0.2cm]
\noindent Typeset by the author in \LaTeX{} using the Latin Modern font family.\\[0.6cm]

\noindent The views and conclusions expressed in this book are those of the author and do not necessarily reflect the positions of any affiliated institution.\\[0.6cm]

\noindent\textbf{Author contact:} Boris Kriuk\\
\vspace*{1cm}
\cleardoublepage

\thispagestyle{empty}
\vspace*{10cm}
\begin{flushright}
    \itshape
    To those who believe\\
    that intelligence is not scale,\\
    but adaptation.
\end{flushright}
\cleardoublepage

\thispagestyle{empty}
\vspace*{8cm}
\epigraph{\itshape ``A system is adaptive not because it learns more,\\
but because it requires less to be told.''}{---\textsc{the author}}
\cleardoublepage

\frontmatter
\pagestyle{plain}
\tableofcontents
\cleardoublepage

\chapter*{Preface}
\addcontentsline{toc}{chapter}{Preface}

This book is an attempt to name something that has been visible in the shadows of artificial intelligence research for more than a decade, but has never quite received its own title. Between the narrow systems we have built and the general intelligence we speculate about, there lies an entire regime of machine behavior that is neither task-bound nor truly autonomous --- a regime in which systems adapt, reconfigure, and select their own structure in response to the problems placed in front of them. I call this regime \emph{Artificial Adaptive Intelligence}, or AAI.

The motivation for this book is not that adaptive methods are new. Meta-learning, AutoML, neural architecture search, continual learning, evolutionary computation, and a growing list of self-tuning techniques have all pushed in this direction for years. The motivation is that these efforts share a common principle that the field has not yet articulated plainly: the steady removal of the human from the loop of parameter specification. Each of these techniques, in its own way, is a step toward systems that need to be told less and infer more.

To make this principle precise, the book adopts a single, deliberately sharp definition:

\begin{quote}
\itshape A system exhibits Artificial Adaptive Intelligence to the extent that it requires no human-specified tunable hyperparameters, while maintaining competitive performance across a diverse distribution of tasks.
\end{quote}

This definition is intentionally uncomfortable. It is falsifiable. It is countable. It refuses the usual refuge of vague appeals to ``flexibility'' or ``robustness.'' And it draws a clean line between four stages of machine intelligence: narrow systems, which expose many hyperparameters to the user; automated systems, which tune those hyperparameters with meta-algorithms that themselves expose hyperparameters; adaptive systems, which expose none; and general systems, which go further still by forming their own goals.

The book is organized around this spine. Each chapter asks the same underlying question --- \emph{which hyperparameter does this technique eliminate, and how?} --- and uses that question to bring together work from aerospace, finance, geophysics, vision, and language modeling that would otherwise sit in disconnected literatures. The ambition is not to claim that zero-hyperparameter systems exist today. They do not. The ambition is to argue that driving the hyperparameter count toward zero is the correct axis of progress between ANI and AGI, and that naming this axis changes what we measure, what we build, and what we call a success.

I have written the book for researchers, graduate students, and practitioners who have felt, as I have, that something is missing from the binary framing of narrow versus general intelligence. Readers familiar with modern machine learning will find the technical chapters self-contained; readers approaching from adjacent fields should find the argumentative chapters accessible without specialized background. Mathematical formalism appears where it earns its place and is deferred where it does not.

A book of this kind is never written alone, even when it carries a single name on the cover. I am grateful to the collaborators, reviewers, and interlocutors whose work and criticism shaped the ideas collected here, and to the reader who is willing to consider that the next stage of artificial intelligence may not arrive through more parameters, but through fewer.

\begin{flushright}
    \textsc{Boris Kriuk}\\
    \textit{2026}
\end{flushright}

\cleardoublepage

\chapter*{How to Read This Book}
\addcontentsline{toc}{chapter}{How to Read This Book}

The book is divided into five parts. Part~I establishes the framing and defends the central definition of adaptivity. Part~II develops the machinery by which adaptive systems derive what would otherwise be hyperparameters from the structure of the data. Part~III turns to architecture --- how systems can select, morph, and evolve their own structural form. Part~IV addresses self-adaptation during training, where the dynamics of learning themselves become objects of adaptation. Part~V closes with the stability, safety, and philosophical questions that any serious proposal in this space must confront.

Readers interested primarily in the conceptual argument may read Parts~I and~V and the opening chapter of each intervening part. Readers interested in technical content may read the book linearly. Readers approaching as practitioners looking for methods to apply may skip directly to Parts~II through~IV, returning to Part~I for the framing when questions of \emph{why} arise.

The central chapter of the book is Chapter~3, \emph{Parametric Minimality}, which carries the load-bearing argument. If any single chapter should be read first, it is that one.

\cleardoublepage

\mainmatter
\newtheorem{definition}{Definition}[chapter]
\newtheorem{principle}[definition]{Principle}
\newtheorem{claim}[definition]{Claim}
\newtheorem{proposition}[definition]{Proposition}

\part{Foundations and Motivation}

\vspace*{2cm}
\noindent\itshape
Before we can argue for a new stage of artificial intelligence, we must
first establish what is missing in the current one. Part~I lays the
groundwork. It begins with the gap between narrow and general
intelligence, and the inadequacy of scale alone as a bridge across it.
It examines the hidden cost of modern machine learning --- the armies
of hyperparameters, the architectural priors, the silent reliance on
the human practitioner --- and prepares the terrain for the central
definition of this book: \emph{adaptivity as the absence of human-specified
tunable parameters}.
\upshape

\vspace*{\fill}
\cleardoublepage

\chapter{Introduction}
\label{chap:introduction}

\epigraph{\itshape ``The greatest obstacle to discovery is not ignorance ---
it is the illusion of knowledge.''}{---\textsc{Daniel J. Boorstin}}

\noindent
The public history of artificial intelligence is often told as a story
of two poles. At one end, the narrow systems: chess engines, speech
recognizers, recommender engines, protein folders, self-driving
perception stacks. At the other, the imagined general intelligence:
a machine that can do, in principle, anything a competent human can do,
and perhaps a great deal more. Between these two poles lies an enormous
conceptual blank. We speak fluently about what narrow intelligence is
and about what general intelligence might be, but we have almost no
language for the territory in between.

This book argues that the territory in between is not empty. It is, in
fact, where most of the interesting research of the past fifteen years
has quietly taken place --- in meta-learning, neural architecture search,
AutoML, continual learning, evolutionary computation, physics-informed
modeling, world models, and a growing family of self-tuning methods. What these
efforts share has not yet been named. Naming it is the purpose of this
book. I will call it \emph{Artificial Adaptive Intelligence}, or AAI,
and I will argue that it is not a minor waypoint on the road from narrow
to general intelligence, but the road itself.

The argument proceeds from a single definition, deliberately sharp:

\begin{definition}[Artificial Adaptive Intelligence]
\label{def:aai}
A system exhibits Artificial Adaptive Intelligence to the extent that
it requires no human-specified tunable hyperparameters while achieving
competitive performance across a diverse distribution of tasks.
\end{definition}

\noindent
The definition is uncomfortable on purpose. It is falsifiable. It is
countable. It refuses refuge in vague appeals to ``flexibility'' or
``robustness.'' Every chapter of this book will return to it, and every
technique discussed will be evaluated by the same question: \emph{which
hyperparameter does it eliminate, and how?}

Before that project can begin, two groundworks must be laid. The first
is an honest account of why the gap between narrow and general
intelligence cannot be closed by scale alone --- a claim which, in the
present moment, runs against a fashionable current. The second is an
honest account of what modern machine learning actually costs, once we
look past its benchmark numbers to the invisible infrastructure of
human tuning, architectural guesswork, and per-task specialization that
keeps the system working. These two sections form the motivation for
everything that follows.

\section{The ANI--AGI Gap and Why ``Scale Alone'' Is an Insufficient Bridge}
\label{sec:ani-agi-gap}

The dominant story of the last five years in artificial intelligence
has been a story of scale. Larger models, trained on more data, with
more compute, have produced capabilities that astonished even their
designers. Capabilities once considered impossible without architectural
innovation --- few-shot reasoning, in-context learning, passable code
generation, cross-domain transfer --- emerged, it appeared, as
byproducts of simply making the same systems bigger. It is therefore
not surprising that a significant portion of the research community,
and nearly all of the industrial community, has converged on a simple
extrapolation: if scale produced these capabilities, more scale will
produce the rest. The gap between narrow and general intelligence, on
this view, is quantitative, not qualitative. It will close on its own,
given enough parameters, tokens, and accelerators.

This book does not dispute the empirical record. The capabilities are
real. The scaling curves are real. What it disputes is the conclusion.
Scale, as currently practiced, is a powerful amplifier of a particular
kind of adaptation --- implicit, gradient-based, and purchased at
enormous computational expense --- but it does not, by itself,
transform narrow intelligence into general intelligence. It transforms
narrow intelligence into \emph{broader} narrow intelligence: systems
that cover more of the space of tasks they were implicitly trained for,
while remaining structurally fixed, hyperparameter-laden, and dependent
on human design choices that scale cannot eliminate.

To see why, it helps to distinguish three properties that any candidate
bridge from ANI to AGI must possess.

\paragraph{Property 1: Coverage.}
The system must handle a wide variety of tasks, including tasks it was
not explicitly trained to solve. Large-scale pretrained models perform
extraordinarily well on this axis. A modern foundation model can write
poetry, summarize law, explain thermodynamics, and debug Python,
often at a level indistinguishable from a competent human generalist.
On coverage, scale is not merely sufficient; it is triumphant.

\paragraph{Property 2: Autonomy of adaptation.}
The system must adapt to a new task without a human manually redesigning
it, retuning it, or rebuilding it. Here the picture is more mixed.
Foundation models adapt to new tasks primarily through prompting,
fine-tuning, and retrieval --- each of which reintroduces, in subtler
forms, the same human-in-the-loop parameter specification that the
field claims to have transcended. Choosing a prompt template is a form
of hyperparameter selection. Choosing a fine-tuning learning rate is a
form of hyperparameter selection. Choosing what to retrieve, when,
and how to weight it is a form of hyperparameter selection. The human
has been moved, not removed.

\paragraph{Property 3: Structural adaptation.}
The system must be able to change its own structure --- its parameter
count, its layer configuration, its inductive biases --- in response
to the structure of the problem it faces. On this axis, scale has
contributed essentially nothing. A 70-billion-parameter transformer
faced with a problem better solved by a decision tree does not become
a decision tree. It remains a transformer, applies its transformer
machinery, and either succeeds or fails within that fixed structural
commitment. The most sophisticated modern models are, in this specific
sense, as rigid as the narrow systems they were meant to supersede.

\medskip
Scale, then, has made enormous progress on the first property, partial
progress on the second, and almost none on the third. A bridge from
ANI to AGI that is strong on coverage but weak on autonomy and
structural adaptation is not a bridge; it is a very long pier. It
extends our reach, but it does not cross the gap.

There is a deeper reason to be skeptical of scale-as-bridge, which has
less to do with engineering and more to do with information. The
central claim of the scaling hypothesis, in its strong form, is that
the structure of general intelligence is latent in the statistical
regularities of sufficiently large and diverse datasets, and that a
sufficiently large parametric model will extract it. This is a bold
empirical bet, and it may in some attenuated form be true. But it has
a conspicuous cost. The amount of information a human designer must
still inject into such a system --- through data curation, objective
specification, reward modeling, safety tuning, evaluation design, and
deployment constraints --- has not decreased as models have grown. If
anything, it has grown with them. The humans required to build, align,
and operate a frontier model in 2026 outnumber those required to build,
align, and operate a narrow model of 2016 by orders of magnitude. The
\emph{system} has grown; the \emph{dependence on human specification}
has not shrunk. On the axis this book cares about --- parametric
minimality --- scale is not making things better. In several important
senses, it is making them worse.

This is the gap that AAI is meant to address. Where scale seeks to make
a single fixed system cover more territory, adaptive intelligence seeks
to make the system itself change shape in response to territory. Where
scale seeks to absorb the hyperparameters of all possible tasks into
the weights of one enormous model, adaptive intelligence seeks to
eliminate the hyperparameters altogether. These are not competing
research programs in the sense that they seek the same destination by
different means. They are competing hypotheses about what the
destination is.

\section{The Hidden Cost of Modern ML: Hyperparameters, Architectural Priors, and Human-in-the-Loop Tuning}
\label{sec:hidden-cost}

Modern machine learning carries a cost that rarely appears in the
headline numbers. The cost is not compute, and it is not data. It is
the quiet, accumulated burden of human specification --- the
hyperparameters, the architectural priors, the training schedules,
the loss weights, the augmentation pipelines, the data-cleaning
heuristics, and the countless small decisions that make the difference
between a model that works and a model that does not. This section
makes that cost visible.

Consider what it takes to bring a modern deep learning system from
conception to deployment. A practitioner must choose, at minimum:

\begin{itemize}
    \item a model family (transformer, convolutional, recurrent, graph, hybrid);
    \item an architectural variant within that family (depth, width, attention pattern, normalization, activation);
    \item an initialization scheme;
    \item an optimizer (SGD, Adam, AdamW, Lion, Shampoo) and its internal constants;
    \item a learning-rate schedule (constant, cosine, warm-up, cyclic);
    \item a batch size and gradient-accumulation strategy;
    \item a regularization regime (weight decay, dropout, label smoothing, stochastic depth);
    \item an augmentation pipeline, if applicable;
    \item a loss function or combination of losses, with weighting coefficients;
    \item an early-stopping criterion;
    \item a checkpoint-selection policy;
    \item an evaluation protocol whose design silently affects all of the above.
\end{itemize}

Each of these decisions is a hyperparameter in the broad sense: a
quantity that the human chooses, that strongly affects the outcome,
and that is not derived from the data by the system itself. The total
dimension of this decision space, for a typical applied project, is
rarely below fifty and sometimes exceeds several hundred. A substantial
fraction of a practitioner's working life is spent searching this space
--- not conducting science, not discovering structure, not advancing
the field, but tuning.

The defense of such state of affairs is usually pragmatic. Default
values work. Community norms converge. A competent practitioner with
good intuitions can shortcut most of the search. All of this is true,
and all of it misses the point. The fact that defaults work is not
evidence that hyperparameters are unimportant; it is evidence that the
community has collectively invested enormous effort in discovering
good defaults for a narrow slice of the task distribution. Change the
slice, and the defaults fail. Move from vision to tabular data, from
English to low-resource languages, from clean benchmarks to noisy
production distributions, from stationary problems to non-stationary
ones, and the defaults collapse. The humans re-enter the loop. The
tuning resumes.

\begin{principle}[The Hyperparameter Tax]
\label{prin:hyperparameter-tax}
The performance of a modern machine learning system on any given task
is the performance of its underlying algorithm \emph{minus} a tax paid
in human hours of hyperparameter selection. This tax is invisible in
benchmark reports, which typically present post-tuning numbers, and it
scales with the distance between the deployment task and the tasks for
which community defaults were developed.
\end{principle}

\noindent
The hyperparameter tax is only the most visible component of the
hidden cost. Beneath it lies a second, deeper layer: the architectural
prior.

Every model family encodes assumptions about the structure of the
problems it is suited for. Convolutional networks assume translation
equivariance and local spatial structure. Transformers assume that
useful relationships can be captured by pairwise attention over tokens.
Recurrent networks assume temporal causality and bounded dependence.
These assumptions are not neutral; they are inductive biases, and they
are chosen by the human practitioner before the system sees a single
data point. When the match between the assumption and the problem is
good, the resulting model is efficient and generalizes well. When the
match is poor, no amount of tuning, scale, or data rescues the system
--- it is solving the wrong problem with the right tools.

The field has absorbed this fact so thoroughly that it no longer
recognizes it as a limitation. We describe the selection of an
architecture as an act of engineering judgment, a matter of
``knowing the domain.'' But from the perspective of adaptive
intelligence, it is nothing of the kind. It is a hyperparameter of
the highest order --- a discrete choice among model families, made by
a human, that commits the system to a particular view of the world
before any data has been examined. A system that selects its own
architecture from the data is doing something qualitatively different
from a system whose architecture is handed to it by a practitioner,
even if both contain identical internal components.

The third layer of hidden cost is the most subtle: human-in-the-loop
tuning that does not appear as tuning. Consider the typical life of a
production ML system. An engineer trains a model. The model goes into
production. Its performance drifts. An alert fires. A different
engineer re-examines the data distribution, notices a shift, adjusts
the preprocessing, retrains the model, re-tunes a handful of
hyperparameters, redeploys, and returns to other work. None of this
activity appears in the model card. None of it is described in the
paper. From the outside, the system appears to be performing
autonomously. From the inside, it is being continuously rebuilt by a
team of humans whose labor has been rendered invisible by the framing
of ``model maintenance.''

The invisibility matters, because it distorts our collective estimate
of how close our systems are to autonomy. A model that requires
quarterly retraining by a team of five engineers is not an autonomous
system with a light maintenance overhead; it is a partially automated
pipeline whose automated portion is shrinking as the non-stationarity
of the environment grows. If we are serious about the progression
from narrow to general intelligence, we must count this labor honestly,
and we must treat its elimination as a research problem of the first
rank.

\begin{claim}[The True Cost of ANI]
\label{claim:true-cost}
The true cost of a narrow artificial intelligence system is not the
cost of training it once, but the cost of the continuous human
specification required to keep it performing as the world around it
changes. Systems that appear narrow and stable from the outside are,
from the inside, cooperative human--machine constructions whose
machine component alone would not survive without the human one.
\end{claim}

\noindent
This is the gap that AAI must close. Not the gap between ANI's
benchmark numbers and AGI's imagined capabilities, but the gap
between the labor currently required to operate an ANI system and the
labor that a truly adaptive system would require --- which, in the
limit, is none. The chapters that follow examine, one at a time, the
specific forms that this labor takes, and the specific techniques
that have begun to eliminate it.

The book's central definition --- that an AAI system is one with no
human-specified tunable hyperparameters --- is offered in this spirit.
It is not a claim that such systems exist today. They do not. It is a
claim about the correct axis along which to measure progress between
the narrow and the general. Every hyperparameter eliminated is a step
along that axis. Every architectural choice deferred to the data is a
step along that axis. Every retraining cycle automated away is a step
along that axis. Where scale measures progress in parameters added,
AAI measures progress in parameters \emph{removed} --- from the
human's responsibility, and folded into the system's own capacity
for adaptation.

This is the argument of the book. Chapter~\ref{chap:introduction} has
stated it. The chapters that follow will defend it.

\section{Thesis Statement: Adaptivity --- Not Scale, Not Generality --- Is the Missing Intermediate Capability}
\label{sec:thesis-statement}

The preceding sections have argued, negatively, that scale is an
insufficient bridge from narrow to general intelligence, and that the
hidden cost of modern machine learning is a continuous stream of
human specification that benchmark numbers conceal. This section
states, positively, what this book proposes in their place.

The thesis of this monograph can be compressed into a single sentence:

\begin{quote}
\itshape
The missing intermediate capability between Artificial Narrow
Intelligence and Artificial General Intelligence is neither more scale
nor more generality, but \textbf{adaptivity}: the capacity of a system
to reconfigure itself --- its parameters, its structure, and its
learning dynamics --- in response to the problem it faces, without a
human re-entering the loop to make those reconfigurations on its
behalf.
\upshape
\end{quote}

\noindent
The claim is deliberately framed as a claim about capability, not
about architecture, ideology, or aesthetics. It does not prescribe a
particular family of models. It does not prescribe symbolic or
subsymbolic methods, gradient-based or evolutionary search, neural
or neuro-symbolic integration. It prescribes a \emph{property} that
the final system must have, and it evaluates candidate methods by
the extent to which they contribute that property.

Three clarifications are in order before the thesis can do any work.

\paragraph{Adaptivity is not flexibility.}
A flexible system is one that can be configured to perform many
different tasks. An adaptive system is one that configures itself.
A pretrained foundation model is flexible in the first sense and
largely not in the second: it can be prompted, fine-tuned, or
adapted to thousands of downstream tasks, but each of those adaptations
requires a human to specify the prompt, the fine-tuning recipe, or the
adapter configuration. Flexibility is a property of the system's
\emph{range}. Adaptivity is a property of the system's \emph{autonomy
within that range}.

\paragraph{Adaptivity is not generality.}
A general system is one that can, in principle, solve any problem
within some broad class. An adaptive system is one that, when placed
in front of a specific problem, modifies itself to solve that specific
problem without external intervention. These properties are related
but distinct. A system can be general but not adaptive --- a fixed,
very large model capable of being prompted into many roles. A system
can be adaptive but not fully general --- a self-tuning solver for
partial differential equations that chooses its own discretization,
step size, and basis but solves nothing outside that domain. The
argument of this book is that generality without adaptivity is
brittle, and that adaptivity without generality is still enormously
valuable, and that the intermediate stage on the way to AGI is
characterized by the latter rather than the former.

\paragraph{Adaptivity is measurable.}
This is the property that distinguishes the thesis of this book from
a philosophical gesture. Adaptivity, as defined here, admits a
quantitative proxy: the number of human-specified tunable
hyperparameters that the system requires per task. A system that
requires fifty such parameters to solve a new task is less adaptive
than one that requires five. A system that requires none is maximally
adaptive, in the sense this book cares about. This metric is crude;
it does not capture every nuance of what adaptation might mean; it can
be gamed by absorbing hyperparameters into training-time constants
that are themselves human-selected. But it is operational, it is
countable, and it points in the right direction. The rest of the book
will refine it.

\begin{principle}[The Adaptivity Thesis]
\label{prin:adaptivity-thesis}
Progress from Artificial Narrow Intelligence to Artificial General
Intelligence proceeds through a distinct intermediate stage,
characterized by the progressive elimination of human-specified
hyperparameters from the learning system. This stage is
\emph{Artificial Adaptive Intelligence}, and its identification as a
stage --- rather than as a collection of miscellaneous engineering
improvements --- is the central claim of this monograph.
\end{principle}

\noindent
The thesis has consequences. If it is correct, then several
contemporary research directions should be reinterpreted. Neural
architecture search is not a niche subfield; it is a direct assault
on the most consequential hyperparameter of all. Meta-learning is
not a specialized technique for few-shot problems; it is the generic
mechanism by which a system learns its own learning dynamics.
Hyperparameter-free optimizers are not optimizer-research trivia;
they are load-bearing infrastructure for any system that aspires to
operate without human supervision. Evolutionary and population-based
methods, long treated as quaint alternatives to gradient descent,
acquire new status as search procedures over structural choices that
gradient descent cannot make. The thesis does not invent these
fields; it unifies them.

The thesis also has implications for what counts as progress. Under
the scaling hypothesis, progress is measured in model size, dataset
size, and benchmark performance. Under the adaptivity thesis,
progress is measured in hyperparameters eliminated, in architectures
auto-discovered, in retraining cycles automated, and in the breadth
of tasks that a single fixed system-design can handle without human
re-intervention. These two metrics can move together --- a larger
model may enable more self-tuning --- but they are not the same, and
conflating them has been a persistent source of confusion in the
field.

\section{Defining Artificial Adaptive Intelligence}
\label{sec:defining-aai}

With the thesis in place, the definition stated in
Section~\ref{sec:ani-agi-gap} can now be elaborated. The definition
is deliberately operational and deliberately narrow; its virtues are
those of a measuring instrument, not of a philosophical exegesis.

\begin{definition}[Artificial Adaptive Intelligence, elaborated]
\label{def:aai-elaborated}
A system $S$ exhibits \emph{Artificial Adaptive Intelligence} with
respect to a task distribution $\mathcal{T}$ to the extent that:
\begin{enumerate}
    \item[(i)] for any task $T \in \mathcal{T}$, $S$ achieves performance competitive with a task-specialized baseline $B_T$;
    \item[(ii)] the human-specified tunable hyperparameters of $S$ do not depend on $T$;
    \item[(iii)] $S$ reconfigures its internal state --- parameters, structure, or learning dynamics --- based on properties of $T$ that $S$ itself infers.
\end{enumerate}
\end{definition}

\noindent
Each clause requires commentary.

\paragraph{Clause (i): competitive performance.}
Adaptivity without performance is not interesting. A system that
requires no human tuning because it performs uniformly poorly
satisfies the letter of the definition and violates its spirit. The
clause therefore requires performance competitive with a
task-specialized baseline --- a system that \emph{was} hand-tuned for
the specific task. The word ``competitive'' is used instead of
``superior'' because the relevant comparison is not whether AAI beats
specialization, but whether it approaches specialization while paying
a far smaller human cost. A system that matches, say, 95\% of the
performance of a hand-tuned baseline across a thousand tasks, using
zero human tuning on any of them, is an enormous scientific achievement
even though it loses every individual comparison.

\paragraph{Clause (ii): task-independent hyperparameters.}
This is the definition's sharpest clause. It does not say the system
must have no hyperparameters at all --- a requirement that would be
vacuous, since any learning rule has constants somewhere. It says
that whatever hyperparameters the system has must be fixed across the
task distribution. They are design-time constants of the
\emph{algorithm}, not deployment-time constants of the
\emph{application}. A learning rate that must be tuned per task
violates this clause. A learning-rate \emph{controller} whose constants
are fixed across tasks, and which itself produces the per-task
learning rate, satisfies it. The clause draws a bright line between
tuning internal to the system and tuning imposed from outside.

\paragraph{Clause (iii): self-inferred reconfiguration.}
The third clause prevents a degenerate reading of the first two.
A system could, in principle, satisfy (i) and (ii) by being a very
large, fixed, universal function approximator that is powerful enough
to absorb the entire task distribution into its weights at training
time, with no per-task adaptation at all. Such a system is not
adaptive; it is pretrained. The third clause requires that the
system, at the moment of encountering a new task, does something
to reconfigure itself on the basis of that task --- whether by
gradient-based fine-tuning, by in-context learning, by structural
search, by hyperparameter inference, or by any other mechanism that
makes the system's behavior on $T$ depend on properties the system
itself has inferred about $T$. Adaptivity, in this sense, is an
active process, not a static capacity.

\medskip
Two properties of the definition are worth emphasizing. The first is
that it admits degrees. A system is not simply adaptive or
non-adaptive; it exhibits AAI \emph{to the extent that} it satisfies
the three clauses. This is important because no contemporary system
satisfies them fully, and a binary definition would therefore declare
AAI non-existent and leave us with nothing to study. A graded
definition lets us measure the current state of the art, track its
movement, and compare candidate methods on a continuous scale.

The second property is that the definition is \emph{relative to a
task distribution}. A system may be maximally adaptive with respect
to one distribution and completely non-adaptive with respect to
another. A self-tuning numerical solver may eliminate every
hyperparameter for the distribution of elliptic PDEs and fail
entirely on hyperbolic ones. This is not a flaw in the definition;
it is a feature. Adaptivity is always adaptivity-for-something, and
one of the central empirical questions of the field is how the set
of distributions over which a system is adaptive grows as the
underlying methods improve. The trajectory from narrow to general
intelligence, viewed through this lens, is the trajectory along
which the task distributions supported by adaptive systems
progressively widen.

\begin{definition}[Adaptivity Index]
\label{def:adaptivity-index}
For a system $S$ and task distribution $\mathcal{T}$, define the
\emph{adaptivity index} $\alpha(S, \mathcal{T}) \in [0,1]$ as
\[
\alpha(S, \mathcal{T}) \;=\; \Bigl(1 - \tfrac{h(S, \mathcal{T})}{h_{\max}(\mathcal{T})}\Bigr)\cdot \rho(S, \mathcal{T}),
\]
where $h(S, \mathcal{T})$ is the expected number of human-specified
tunable hyperparameters required to deploy $S$ on a task drawn from
$\mathcal{T}$, $h_{\max}(\mathcal{T})$ is the corresponding count for
a fully hand-tuned task-specialized baseline, and
$\rho(S, \mathcal{T}) \in [0,1]$ is the performance ratio of $S$ to
that baseline, averaged over $\mathcal{T}$.
\end{definition}

\noindent
The adaptivity index is intentionally crude. It collapses a great deal
of structure into two numbers. Its purpose is not to adjudicate
marginal disputes between methods but to make the argument of this
book \emph{numerical}. A method that increases $\alpha$ is making
progress toward AAI. A method that leaves $\alpha$ unchanged, however
impressive its benchmark numbers, is not --- or at least, not on the
axis this book cares about. The chapters that follow will use this
index, in various refined forms, as a common evaluative vocabulary.

\section{Scope, Contributions, and Roadmap of the Monograph}
\label{sec:scope-roadmap}

Having stated the thesis and fixed the definition, the remainder of
this section describes what this book will and will not attempt, what
it claims to contribute, and how its argument unfolds across the
chapters that follow.

\subsection*{Scope}

The scope of this monograph is defined by three commitments.

First, the book concerns itself with \emph{machine} adaptivity, not
biological adaptivity. It will draw on biological analogies where they
illuminate engineering choices --- evolution, development, neural
plasticity, immune adaptation --- but it will not attempt to survey
the biology in its own right, and it will not treat biological
fidelity as a design criterion. A method is interesting to this book
insofar as it contributes to a measurable reduction in the adaptivity
index, regardless of whether it resembles anything nature has
produced.

Second, the book concerns itself with \emph{learning} adaptivity, not
\emph{behavioral} adaptivity. A reinforcement-learning agent that
adapts its behavior within an episode is a familiar and well-studied
object; the behavioral literature is vast, and this book does not
attempt to absorb it. What this book is about is the adaptation of
the learning system \emph{itself} --- its architecture, its
optimizer, its hyperparameters, its structural commitments --- across
tasks and across time. Behavioral adaptation appears in these pages
only where it intersects with learning-system adaptation, as in
meta-reinforcement learning or continual control.

Third, the book is \emph{theoretical and integrative}, not
experimental. It does not report new benchmark results. It does not
propose new algorithms in the sense of offering implementations that
compete on leaderboards. What it offers is a conceptual framework, a
vocabulary, a set of definitions, and a synthesis of existing work
under those definitions. Its contribution, if any, is to make visible
a field that is currently distributed across a dozen subcommunities
that do not recognize themselves as parts of a single enterprise.

\subsection*{Contributions}

The contributions of the monograph are five.

\begin{enumerate}
    \item \textbf{A name and a definition.} The book names and operationally defines Artificial Adaptive Intelligence as a distinct stage between narrow and general intelligence, characterized by the progressive elimination of human-specified hyperparameters.
    \item \textbf{A unifying framework.} The book places meta-learning, neural architecture search, AutoML, continual learning, self-supervised learning, physics-informed models, and evolutionary methods within a single framework organized by the adaptivity index.
    \item \textbf{A critique of the scaling hypothesis.} The book offers a structured argument that scale, while powerful, addresses only one of the three properties (coverage, autonomy, structural adaptation) required to bridge the ANI--AGI gap, and that the remaining two are the proper domain of AAI.
    \item \textbf{An accounting of hidden costs.} The book makes explicit the human-in-the-loop costs of current machine learning practice and argues that their elimination, not merely their amortization across ever-larger models, is the correct measure of progress.
    \item \textbf{A research agenda.} The book closes with a set of concrete open problems whose solutions would measurably advance the adaptivity index and thereby the field.
\end{enumerate}

\subsection*{Roadmap}

The monograph is organized in five parts.

\textbf{Part I} --- of which the present chapter is the opening ---
lays the conceptual groundwork. It establishes the ANI--AGI gap, the
insufficiency of scale, the hidden costs of modern ML, and the
thesis that adaptivity is the missing intermediate capability. It
closes with a formal development of the adaptivity index and a
survey of the historical precedents from which the AAI framing
draws.

\textbf{Part II} examines \emph{parametric adaptivity}: the
elimination of hyperparameters that govern the learning process
itself. Chapters in this part treat hyperparameter-free optimizers,
learned learning-rate schedules, meta-learned update rules, and the
broader meta-learning program. The unifying question is: can the
system choose its own learning dynamics?

\textbf{Part III} examines \emph{structural adaptivity}: the
elimination of hyperparameters that govern the model's form. Chapters
in this part treat neural architecture search, evolutionary structural
search, dynamic and growing networks, and modular systems whose
composition is determined at runtime. The unifying question is: can
the system choose its own shape?

\textbf{Part IV} examines \emph{distributional and temporal
adaptivity}: the elimination of the human maintenance loop. Chapters
in this part treat continual learning, non-stationarity, self-supervised
adaptation to distribution shift, and test-time training. The
unifying question is: can the system keep itself working as the
world changes?

\textbf{Part V} consolidates. It returns to the thesis, re-evaluates
it in light of the techniques surveyed, discusses the limits of the
adaptivity index, considers safety and governance implications of
systems that tune themselves, and closes with a research agenda.

Each chapter follows a common pattern. It identifies a specific
hyperparameter or class of hyperparameters. It surveys the methods
that have been proposed to eliminate it. It evaluates those methods
against the adaptivity index. And it closes with a discussion of
what remains --- what residue of human specification the best
current methods still require, and what would be needed to remove
it.

The book can be read linearly or sampled. Readers interested in the
philosophical and conceptual case for AAI may find Parts~I and~V
sufficient. Readers interested in the technical substance may begin
with Parts~II, III, or~IV according to their interests; each is
self-contained, though cross-references are provided where an
argument in one part depends on a result in another. Readers seeking
a critical position from which to evaluate the current moment in
artificial intelligence --- the scale-first, foundation-model
moment --- may find the juxtaposition of Chapter~\ref{chap:introduction}
and the concluding chapters most useful.

\medskip
The remainder of this part develops the historical and conceptual
background against which the arguments of Parts~II--IV will be made.
The next chapter turns to that history: the forgotten prehistory of
adaptive systems in cybernetics, the rise and retreat of hyperparameter
sensitivity as a research concern, and the specific moments at which
the field came close to the AAI framing before turning away from it.

\cleardoublepage

\chapter{A Taxonomy of Intelligence Paradigms}
\label{chap:taxonomy}

\epigraph{%
\itshape
``The classification of the constituents of a chaos, nothing less is
here essayed.''%
}{--- Herman Melville, \textit{Moby-Dick}}

\noindent
The previous chapter argued that a stage of artificial intelligence
is missing from the field's self-description: a stage between the
narrow systems that dominate deployment and the general systems that
dominate discourse. Arguing that a stage is missing, however,
requires a stable account of the stages it sits between. Without
such an account, the claim that something is missing degenerates
into a claim that something is merely under-theorized, and the
proposal to fill the gap becomes a proposal to redraw arbitrary
lines.

This chapter therefore undertakes a task that is, at first glance,
unglamorous: it re-examines the standard taxonomy of
\emph{narrow}, \emph{general}, and --- in some versions ---
\emph{super} intelligence, and asks what this taxonomy actually
carves. It argues that the customary three-way division is
structurally incomplete, that its endpoints are better understood
than is usually acknowledged while its middle is worse understood
than is usually acknowledged, and that the missing middle is the
object of this book.

The chapter proceeds as follows. Section~\ref{sec:ani}
characterizes Artificial Narrow Intelligence as it is actually
practiced, distinguishing its genuine strengths from a brittleness
and tuning burden that the benchmark literature systematically
undercounts. Section~\ref{sec:agi-premature} turns to Artificial
General Intelligence, surveying the major definitions in the
literature, distinguishing the empirical questions from the
aspirational ones, and arguing that direct pursuit of AGI ---
whatever one means by the term --- is methodologically premature
given the state of the intermediate capabilities on which it would
have to rest.

Subsequent sections of the chapter, developed after those below,
introduce Artificial Adaptive Intelligence as the missing stage, lay
out its defining capabilities, and compare it systematically with
adjacent but distinct proposals (AutoML, meta-learning,
continual learning, foundation models).

\section{Artificial Narrow Intelligence: Strengths, Brittleness, and the Tuning Burden}
\label{sec:ani}

Artificial Narrow Intelligence is the only form of artificial
intelligence that has ever been deployed at scale. Every production
system --- every recommender, every translator, every classifier,
every speech recognizer, every large language model serving
requests --- is a narrow system in the operational sense developed
below, irrespective of whether its underlying model is small or
large, specialized or foundation, hand-engineered or learned. This
framing may strike the reader as provocative, since the same large
language models are routinely described as \emph{general} by their
developers and their users. The apparent tension dissolves once the
term \emph{narrow} is given operational content.

\subsection{What Narrowness Actually Means}

The standard textbook definition of ANI is that it is intelligence
\emph{confined to a single task or domain}. This definition, though
widely repeated, is inadequate for contemporary systems. A modern
language model is not confined to a single task; it translates,
summarizes, answers questions, writes code, and performs hundreds of
other linguistic operations within a single serving endpoint.
Declaring such a system narrow because it ``only does language''
stretches the notion of a single domain to the point of vacuity, and
declaring it general because it does many things inside that domain
erases the distinction the taxonomy is meant to preserve.

A more useful definition shifts the locus of narrowness from the
\emph{scope of tasks} to the \emph{locus of adaptation}:

\begin{definition}[Artificial Narrow Intelligence, operational]
\label{def:ani}
A system $S$ is \emph{narrow} with respect to a deployment context
$\mathcal{C}$ if, for any change in $\mathcal{C}$ that requires a
change in $S$'s behavior, the required change in $S$ is produced by
a human acting on $S$ from outside, rather than by $S$ acting on
itself from within.
\end{definition}

\noindent
On this definition, narrowness is not a property of what a system
can do; it is a property of who does the adapting when the context
changes. A specialist image classifier trained on one domain and
unable to generalize to another is narrow in this sense, but so is
a foundation model that must be re-prompted, re-fine-tuned, or
re-trained whenever the deployment context shifts in a way that
matters. Both systems respond to context change through human
intervention. That is what makes them narrow. The difference
between them is in the \emph{breadth} of tasks each can be humanly
tuned to, not in the \emph{autonomy} with which each adapts.

This definition has three consequences that will matter throughout
the book.

\paragraph{Consequence 1: Narrowness is an adaptive property, not a capability property.}
It is possible for a very capable system to be very narrow, and for
a modestly capable system to be less narrow. Capability and
narrowness are independent axes. The scaling-hypothesis literature
tends to collapse them, reading rising capability as declining
narrowness; the present framework does not.

\paragraph{Consequence 2: Foundation models are (mostly) narrow.}
A pretrained model that is adapted to each new context by a human
prompt-engineer, fine-tuner, or retrieval-pipeline author is narrow
in the operational sense. The human is the adaptation mechanism.
The fact that adaptation is now called ``prompting'' instead of
``training'' does not change its human-in-the-loop character. The
industrial apparatus of prompt engineering, retrieval augmentation,
evaluation harnesses, and continuous fine-tuning is, by this
definition, the visible evidence of narrowness, not its refutation.

\paragraph{Consequence 3: Narrowness admits degree.}
A system that requires a hundred human interventions per deployment
is more narrow than one that requires ten. The ANI-to-AAI transition
is the transition along this degree, measured by the adaptivity
index of Definition~\ref{def:adaptivity-index}.

\subsection{The Genuine Strengths of ANI}

It is easy, in a book arguing for a new paradigm, to caricature the
old one. ANI deserves more careful treatment. Its strengths are
real, and any proposal to supersede it must either preserve those
strengths or explain what is gained by their sacrifice.

The first genuine strength of ANI is \emph{predictable performance
on in-distribution inputs}. A well-tuned narrow system, evaluated on
inputs drawn from the same distribution as its training data, behaves
in a way that is statistically characterizable, auditable, and
certifiable. Whole regulatory regimes --- in medicine, finance,
transportation --- depend on the ability to bound a system's error
rate on a defined distribution. Narrow systems support such bounds
because their behavior is fixed once their human-tuned configuration
is fixed.

The second strength is \emph{compositional tractability}. A narrow
system has a well-defined interface, well-defined inputs and outputs,
and a well-defined behavior within those boundaries. It can be
composed with other narrow systems, tested in isolation, and
replaced by an improved version without a redesign of the pipeline
that contains it. Software engineering practice is built around
components with this character. Adaptive systems, by their nature,
threaten compositional tractability --- a system that changes its
own behavior is a system whose interface is, in some sense, not
fixed --- and any successor to ANI must engage with this cost
directly.

The third strength is \emph{efficient specialization}. For a task
that is truly stationary, truly narrow, and truly well-characterized,
a hand-tuned specialist will outperform, per unit compute, any
general-purpose system. The cost of generality is paid in parameters,
in data, and in inference latency. For applications where the task
distribution is narrow enough to justify specialization, this cost
is wasteful, and ANI is the correct design choice. The argument of
this book is not that ANI should be abandoned; it is that the
\emph{field as a whole} should not be satisfied with ANI as its
frontier.

\subsection{The Brittleness Problem}

The weakness of ANI, widely acknowledged, is brittleness:
degradation of performance when inputs depart from the training
distribution. The literature on this phenomenon is vast, and no
survey is attempted here. Four forms of departure are worth
distinguishing, because they motivate four distinct responses and
call for four distinct capabilities in any successor system.

\begin{enumerate}
    \item \textbf{Covariate shift}: the input distribution changes, while the conditional distribution of outputs given inputs is stable. The system sees unfamiliar inputs but the underlying task is unchanged.
    \item \textbf{Label shift}: the marginal distribution of outputs changes, while the conditional distribution of inputs given outputs is stable. The same phenomena occur, but with different base rates.
    \item \textbf{Concept drift}: the conditional relationship between inputs and outputs itself changes over time. The task the system was trained to perform is no longer the task being asked of it.
    \item \textbf{Adversarial shift}: inputs are chosen by an agent whose objective is to cause the system to err. Departures from the training distribution are not incidental but designed.
\end{enumerate}

\noindent
A narrow system, by Definition~\ref{def:ani}, responds to each of
these only through human intervention. Covariate shift is detected
by a human monitoring dashboards; label shift is corrected by a
human retraining on new data; concept drift is diagnosed by a human
noticing that errors have a pattern; adversarial shift is
countered by a human analyzing the attack. Each of these human
interventions is expensive, slow, and fallible, and each is, in the
framework of this book, a hyperparameter the narrow system has
failed to absorb into itself.

The brittleness problem is therefore not a collection of incidents
to be patched; it is the structural consequence of locating the
adaptation mechanism outside the system. Scaling the system does not
relocate the mechanism. A larger narrow system is still a narrow
system; it is brittle in the same structural sense, though it may
be brittle in fewer places or less visibly.

\subsection{The Tuning Burden}

The complement of brittleness is the tuning burden: the human effort
required to get a narrow system to perform well in the first place,
and to keep it performing well over time. The tuning burden is rarely
quantified in research papers, because research papers report the
performance achieved by the final, tuned configuration, not the
effort expended to discover that configuration. But the effort is
real, and it has been growing, not shrinking, even as the field has
matured.

Consider what must be chosen to deploy a modern learning system.
The architecture --- its family, its depth, its width, its connectivity,
its attention pattern, its normalization scheme. The optimizer ---
its family, its learning rate, its schedule, its warmup, its decay,
its momentum, its weight decay, its gradient clipping. The training
regime --- its batch size, its data ordering, its mixing of sources,
its curriculum, its augmentations, its regularizations, its
stopping criterion. The deployment pipeline --- its quantization, its
serving batch size, its retrieval configuration, its safety filters,
its prompt templates, its evaluation suite. A full enumeration of
the configuration choices for a single modern system runs into the
hundreds. Each of these is, in principle, a hyperparameter; each is,
in practice, either set by default (inheriting someone else's tuning
decision) or set by search (consuming compute that is invisible in
the final paper).

\begin{principle}[The Conservation of Tuning Effort]
\label{prin:conservation-tuning}
In a system that is not itself adaptive, decisions not made by a
human tuner are made by a default, and decisions not made by a
default are made by search. No decision is truly absent. The
question is not whether tuning is performed but where the cost of
tuning is booked.
\end{principle}

\noindent
This principle is not original to this book --- it is implicit in
every honest account of machine learning practice --- but stating it
explicitly clarifies what a shift away from ANI actually requires.
It is not enough to remove human tuners from the loop while leaving
defaults and search procedures unchanged; the defaults and search
procedures are themselves human artifacts, and the adaptivity index
of Definition~\ref{def:adaptivity-index} counts them as such. A
system is adaptive only to the degree that its tuning decisions are
made \emph{by the system, on the basis of the problem}, rather than
\emph{by a human, on the basis of prior experience with adjacent
problems}.

The tuning burden is the cost that ANI pays for its strengths.
Predictable performance, compositional tractability, and efficient
specialization are all purchased by fixing the system's
configuration, and that fixing is done by humans. The burden is
tolerable when tasks are few, stationary, and long-lived, because
the one-time cost of tuning amortizes over a long deployment. It
becomes intolerable when tasks are many, non-stationary, and
short-lived --- which is the regime that modern deployment
increasingly inhabits.

\section{Artificial General Intelligence: Definitions, Open Problems, and Why Direct Pursuit Is Premature}
\label{sec:agi-premature}

If Artificial Narrow Intelligence suffers from too many operational
definitions and too little theoretical unification, Artificial General
Intelligence suffers from the opposite malady. It has been the
subject of more definitions than any other concept in the field, and
yet fewer of those definitions admit operational content than one
might hope. Much of the confusion in contemporary discourse about
AGI --- whether it has been achieved, whether it is near, whether it
is possible --- traces to the fact that different participants in the
discussion use different definitions without marking the difference.

This section does not attempt to adjudicate which definition is
correct. It surveys the main families of definitions, identifies the
empirical commitments each makes, and argues that however the term
is defined, the direct pursuit of AGI is premature given the state
of the intermediate capabilities that any plausible AGI would have
to rest on. The argument is not anti-AGI; it is anti-shortcut.

\subsection{Four Families of Definitions}

The definitions of AGI in circulation can be organized into four
families, each of which captures something genuine and each of which,
taken alone, is incomplete.

\paragraph{(i) Task-breadth definitions.}
These define AGI as a system capable of performing any intellectual
task that a human can perform, at a level comparable to a competent
human. The definition is attributable in various forms to early
proposals for human-level machine intelligence, and it is the
definition most commonly invoked in popular discussion. Its virtue
is that it refers to an observable property: one can, in principle,
enumerate tasks and measure performance on each. Its weakness is
that the enumeration is open-ended, that ``comparable to a competent
human'' is not well-defined across tasks, and that very large
collections of narrow capabilities can approximate the definition
without satisfying whatever the definition was meant to capture.

\paragraph{(ii) Task-transfer definitions.}
These define AGI as a system capable of \emph{learning} any
intellectual task a human can learn, given a reasonable amount of
experience. The emphasis shifts from capability to acquisition: not
``can it do X'' but ``can it learn to do X.'' This family is
conceptually closer to the concerns of this book, because it
foregrounds adaptation. Its weakness is that ``learning'' is itself
ambiguous --- does in-context prompting count? does fine-tuning?
does full retraining? --- and without a principled account of what
learning requires, the definition can be satisfied by systems that
merely simulate learning through sufficiently broad pretraining.

\paragraph{(iii) Capability-ensemble definitions.}
These define AGI by enumerating capabilities that a system must
exhibit: reasoning, planning, memory, perception, action, language,
social cognition, metacognition. The definition is useful for
diagnostic purposes --- given a candidate system, one can check which
capabilities are present --- but it is vulnerable to the charge that
an ensemble of narrow capabilities is not obviously the same thing as
general intelligence, and that the list is never complete.

\paragraph{(iv) Unifying-mechanism definitions.}
These define AGI by the presence of a single underlying mechanism ---
universal inductive inference, a world model, a self-improving
learner, a reflective agent --- from which the observable
capabilities emerge. These definitions are theoretically attractive
and empirically fragile; the proposed mechanisms are often either
uncomputable, unverifiable, or so abstract as to be unfalsifiable.
Their role is mostly inspirational: they point toward what one might
hope an AGI would be, while leaving the engineering question
untouched.

\medskip
No definition in any of these families is operationally satisfactory
on its own. Task-breadth definitions count outputs without counting
costs; task-transfer definitions invoke learning without defining
its mechanism; capability-ensemble definitions are lists; unifying-
mechanism definitions are aspirations. A working definition of AGI
would need to integrate across families, which the field has not
done.

This book does not attempt that integration. It takes a different
route: it argues that regardless of how AGI is eventually defined,
the path to it passes through a capability --- adaptivity --- that
is under-developed relative to the definitions' demands, and that
developing that capability is the correct next step whether the
destination turns out to be AGI in one sense or another.

\subsection{Open Problems That Precede AGI}

To see why direct pursuit of AGI is premature, it helps to list
problems whose resolution any plausible AGI would presuppose, and to
observe the state of each. The list below is not exhaustive; it is
chosen to include problems from across the definitional families,
and to include only problems on which the current state of the art is
demonstrably partial.

\begin{enumerate}
    \item \textbf{Learning from limited data.} A plausible AGI must learn tasks from quantities of experience comparable to what a human requires. Contemporary systems require, on most tasks, experience several orders of magnitude greater. The gap is narrowing in some domains and widening in others, and there is no consensus theory of what closes it.
    \item \textbf{Out-of-distribution generalization.} A plausible AGI must generalize to situations unlike those in its training data. Contemporary systems generalize well within distribution and degrade unpredictably outside it. The characterization of when out-of-distribution generalization will succeed remains open.
    \item \textbf{Compositional reasoning.} A plausible AGI must combine known concepts into novel structures and reason over the combinations. Contemporary systems exhibit partial compositional behavior that is difficult to distinguish from memorized pattern completion, and the conditions under which true composition arises are disputed.
    \item \textbf{Continual learning without catastrophic forgetting.} A plausible AGI must acquire new knowledge without destroying old knowledge. Contemporary systems, trained on sequences of tasks, typically forget earlier tasks as they learn later ones, and the known mitigations remain partial.
    \item \textbf{Self-evaluation and metacognition.} A plausible AGI must know what it knows and, equally, know what it does not know. Contemporary systems are systematically miscalibrated: they produce confident outputs on inputs they cannot reliably handle, and the miscalibration is not reliably removed by scale.
    \item \textbf{Goal inference and specification.} A plausible AGI must infer the goals it is asked to pursue from instructions that are, in practice, under-specified. Contemporary systems follow instructions literally or are tuned by humans toward a target interpretation; they do not robustly infer what the instruction-giver meant.
    \item \textbf{Autonomous adaptation.} A plausible AGI must, when its environment changes, change itself. Contemporary systems adapt through humans in the loop, by Definition~\ref{def:ani}.
\end{enumerate}

\noindent
The list includes problems from the learning-theoretic tradition
(items 1--4), the cognitive-scientific tradition (items 3, 5), the
alignment tradition (item 6), and the engineering tradition (items
4, 7). It spans the definitional families of AGI. And it is, in
every case, a list of problems that would have to be substantially
resolved before any credible claim of AGI could be made --- problems
for which current solutions are, by the admission of the people
working on them, incomplete.

\subsection{Why Direct Pursuit Is Premature}

Given the list above, the argument for the prematurity of direct AGI
pursuit follows in three steps.

\paragraph{Step 1: The open problems are largely adaptivity problems.}
Inspection of the list shows that most of the open problems, when
examined closely, reduce to instances of the inability of
contemporary systems to adapt without human intervention.
Out-of-distribution generalization is the failure to adapt to a new
distribution. Continual learning is the failure to adapt to a new
task without discarding old ones. Metacognition is the failure to
adapt confidence to competence. Goal inference is the failure to
adapt the interpretation of an instruction to the context of its
utterance. Autonomous adaptation is literal. The items that do not
reduce cleanly to adaptivity (compositional reasoning, sample
efficiency) are either enabled by adaptivity or closely entangled
with it. The AGI research program, viewed through the lens of its own
open problems, \emph{is} an adaptivity research program that has not
yet recognized itself as such.

\paragraph{Step 2: The adaptivity problems have tractable near-term targets.}
Unlike the definitional problem of AGI --- which admits no clear
empirical resolution --- the adaptivity problems admit measurable
progress. The adaptivity index of Definition~\ref{def:adaptivity-index}
can be computed for a candidate system on a task distribution. Methods
that raise it can be identified, compared, and combined. Failures
can be localized to particular hyperparameters that the system has
not yet absorbed. This is what a research program looks like when it
has tractable near-term targets, and what AGI, as currently
discussed, does not.

\paragraph{Step 3: Direct pursuit skips the stage on which it depends.}
A research program that aims directly at AGI --- that asks, of each
proposed method, whether it brings us closer to human-level general
intelligence --- must either solve the adaptivity problems as a
by-product or build a system that does not require them solved.
The first alternative is indirect adaptivity research under another
name, and proceeds more efficiently when named. The second alternative
is the scaling hypothesis considered in Chapter~\ref{chap:introduction},
whose difficulties were examined there. In either case, the direct
framing adds no research leverage over a framing that names the
intermediate capability explicitly and develops it on its own terms.

\medskip
The argument is not that AGI is impossible, uninteresting, or
unworthy of eventual pursuit. It is that the current state of the
art does not yet support direct pursuit, that the intermediate stage
of adaptivity is the stage at which genuine progress can be made,
and that progress at that stage is, not coincidentally, progress on
precisely the problems that any future AGI would have to have
solved. Adaptivity is not a detour around AGI; it is the road.

The chapter continues, in the sections that follow, by introducing
Artificial Adaptive Intelligence as the paradigm that names this
road, and by situating it among the adjacent research programs ---
AutoML, meta-learning, continual learning, foundation models --- that
have addressed parts of it without naming the whole.

\section{Positioning AAI: Operational Definition and Distinguishing Criteria}
\label{sec:positioning-aai}

The preceding sections argued that the customary taxonomy of
artificial intelligence has a structural gap: between narrow systems
whose adaptation is performed by humans acting from outside, and
general systems whose definition remains contested and whose direct
pursuit is premature, there is an intermediate stage whose
defining property is that adaptation is performed by the system
acting on itself from within. This section names that stage and
gives it operational content.

The name \emph{Artificial Adaptive Intelligence} is not new; variants
of the term have appeared in the literature on adaptive control,
on evolutionary computation, and on organizational cybernetics,
often referring to quite different things. The present usage is
narrower than some and broader than others, and no claim of
terminological priority is made. What matters is not the label but
the object the label picks out, and the criteria by which systems
are admitted to the category.

\subsection{An Operational Definition}

\begin{definition}[Artificial Adaptive Intelligence, operational]
\label{def:aai-operational}
A system $S$ exhibits \emph{Artificial Adaptive Intelligence} with
respect to a task distribution $\mathcal{T}$ and a context space
$\mathcal{C}$ if, for a substantive fraction of the decisions that
govern $S$'s behavior on tasks drawn from $\mathcal{T}$ in contexts
drawn from $\mathcal{C}$, the decision is:
\begin{enumerate}
    \item made by $S$ itself, on the basis of properties of the
    task and context, rather than by a human acting on $S$ from
    outside;
    \item revisable by $S$ when the task or context changes,
    without requiring retraining, redeployment, or re-specification
    by a human; and
    \item stable under composition, in the sense that $S$'s
    adaptive decisions on one task do not destroy its competence on
    previously encountered tasks within $\mathcal{T}$.
\end{enumerate}
\end{definition}

\noindent
The definition is deliberately graded rather than binary. The phrase
``a substantive fraction of the decisions'' admits that no real system
will absorb all of its configuration into itself, and that adaptivity
is a spectrum along which systems can be placed rather than a line
they can cross. The three clauses correspond, respectively, to
\emph{autonomy} (the system, not a human, makes the decision),
\emph{reversibility} (the decision can be unmade when circumstances
change), and \emph{compositional stability} (adaptation on one task
does not pay for itself by forgetting others).

Each clause is necessary. A system that makes its own decisions but
cannot revise them is merely a system with an unusual compilation
step. A system that revises its decisions but loses prior competence
in the process is a system that cannot be trusted to carry
capabilities forward. A system that preserves prior competence but
relies on human decision-making at the point of adaptation is an
ANI system with good maintenance practice. Only the conjunction
captures what is distinctive about adaptivity.

\subsection{The Adaptivity Index, Restated}

The definition of AAI refers to ``a substantive fraction of
decisions.'' To make this quantitative, recall the adaptivity index
introduced informally in Chapter~\ref{chap:introduction} and now
restated in operational form.

\begin{definition}[Adaptivity Index, operational form]
\label{def:adaptivity-index-operational}
Let $\mathcal{H}(S)$ denote the set of \emph{hyperparameters} of
system $S$ --- the decisions whose values are fixed before $S$
begins operating on any given task and which, if set differently,
would yield measurably different behavior. For each
$h \in \mathcal{H}(S)$, let $\alpha(h) \in [0,1]$ denote the degree
to which $h$ is set by $S$ itself on the basis of the task at hand,
with $\alpha(h)=0$ indicating that $h$ is set by a human (or by a
human-chosen default) and $\alpha(h)=1$ indicating that $h$ is set
by $S$ through task-conditioned inference. The \emph{adaptivity
index} of $S$ is
\[
\mathcal{A}(S) \;=\; \frac{1}{|\mathcal{H}(S)|} \sum_{h \in \mathcal{H}(S)} w(h)\, \alpha(h),
\]
where $w(h) \geq 0$ is a weight reflecting the operational
significance of $h$, normalized so that $\sum_h w(h) = |\mathcal{H}(S)|$.
\end{definition}

\noindent
The index has three properties worth noting. First, it is defined
relative to a stated hyperparameter set; the same system can have
different indices under different choices of $\mathcal{H}$, and
honest reporting requires the set to be specified. Second, it is
weighted, so that absorbing the learning-rate schedule into the
system counts for more than absorbing, say, the default random seed.
Third, the values $\alpha(h)$ are themselves estimates, and the
index is useful chiefly for comparison across systems or versions
rather than as an absolute measurement. A candidate AAI system
should exhibit a substantially higher index than its ANI
predecessor on the same task distribution, and should do so without
degrading the performance that the ANI predecessor achieved. The
absolute value of the index matters less than the direction and
stability of its change.

\subsection{Four Distinguishing Criteria}

Beyond the adaptivity index, four criteria distinguish AAI systems
from neighboring categories. These are used throughout the book to
test whether a candidate system belongs in the category.

\paragraph{Criterion 1: Task-conditioned self-configuration.}
The system adjusts its own configuration --- architecture,
optimizer, regularization, inference strategy --- on the basis of
properties of the task it is presented with, rather than executing
a fixed configuration chosen in advance. The key word is
\emph{conditioned}: it is not enough for the system to have many
possible configurations if the choice among them is made by an
external controller.

\paragraph{Criterion 2: Regime detection.}
The system recognizes, and responds to, qualitative changes in its
operating regime --- shifts in data distribution, task semantics,
resource availability, or feedback signal --- without being told
that such a shift has occurred. Regime detection is the perceptual
side of adaptation; without it, self-configuration is applied
blindly.

\paragraph{Criterion 3: Structural minimality.}
The system exposes only those hyperparameters that cannot, given
current understanding, be absorbed into task-conditioned inference.
Every remaining hyperparameter is an admission of incomplete
adaptivity and is marked as such. This criterion operationalizes
the idea, pursued across the book, that adaptivity is measured not
by what the system does but by what it \emph{no longer asks its
user to decide}.

\paragraph{Criterion 4: Preservation under revision.}
Adaptations that improve performance on the current task do not
degrade performance on previously encountered tasks within the
system's scope, except by an amount the system can bound and
report. This criterion distinguishes adaptation from drift, and
its satisfaction requires the continual-learning machinery discussed
in Section~\ref{sec:related-concepts}.

\medskip
A system that satisfies all four criteria, on a non-trivial task
distribution, is an AAI system in the strong sense. A system that
satisfies some but not others occupies an intermediate position
that is still useful to name. Most contemporary systems satisfy
fragments of Criterion~1 under controlled conditions and little
else; this is the state of the art, and it is the starting point
from which the chapters that follow proceed.

\subsection{What AAI Is Not}

Three clarifications are useful before moving on, since each
corresponds to a confusion that the literature has not consistently
avoided.

\paragraph{AAI is not AGI at a smaller scale.}
An AAI system need not perform arbitrary intellectual tasks; it
need only perform, with autonomous adaptation, the tasks within its
scope. The distinction is between breadth of task and autonomy of
adaptation. AGI, under most definitions, requires both;
AAI requires the latter and bounds the former.

\paragraph{AAI is not automation of the tuning pipeline.}
A pipeline that runs hyperparameter search automatically, without
human attendance, is not thereby an AAI system. The search
procedure and its configuration are themselves human artifacts,
and the adaptivity index counts them as such. AAI requires that the
system's adaptive decisions be made \emph{on the basis of task
properties inferred at use time}, not \emph{on the basis of fixed
search procedures specified in advance}. 

\paragraph{AAI is not online learning plus marketing.}
Systems that update their parameters on arriving data are not
thereby adaptive in the sense of Definition~\ref{def:aai-operational},
unless the update mechanism is itself task-conditioned, reversible,
and preservation-respecting. Many deployed online-learning systems
satisfy none of these; they are ANI systems with a continuous
tuning feed, and they exhibit the pathologies of both ANI
(brittleness to regime shift) and online updating (forgetting of
earlier regimes).

With the operational content of AAI fixed, we can now situate the
paradigm among the adjacent research programs it extends and, in
some cases, subsumes.

\section{Related Concepts: Meta-Learning, AutoML, Continual Learning, Neural Architecture Search}
\label{sec:related-concepts}

AAI, as defined in the previous section, is not a field created
from nothing. It names and integrates a set of capabilities that
have each been developed within their own research programs, and
its novelty lies less in any single capability than in the
insistence that these capabilities belong together and must be
evaluated together. This section reviews the four most directly
relevant programs --- meta-learning, AutoML, continual learning,
and neural architecture search --- and argues, in each case, that
AAI both \emph{subsumes} the program (treating its outputs as
components) and \emph{extends} it (requiring of those components
behaviors the program itself does not guarantee).

\subsection{Meta-Learning}

Meta-learning, or ``learning to learn,'' concerns systems that
improve their ability to acquire new tasks by training on a
distribution of related tasks. The canonical formulation treats
each task as a single example in a higher-level learning problem,
and seeks a learner --- a model, an initialization, an optimizer, a
representation --- that generalizes across the task distribution.
The program has produced substantive results: gradient-based
meta-learners that adapt to new tasks with few examples,
attention-based learners that compose demonstrations at inference
time, and a rich theoretical literature on the conditions under
which meta-learning accelerates acquisition.

\paragraph{What AAI takes from meta-learning.}
The mechanism of task-conditioned adaptation is, in large part, the
mechanism meta-learning has made operational. The idea that a
system can expose an interface by which new tasks are communicated
in small numbers of examples, and that the system's behavior
adjusts accordingly without re-training from scratch, is the
meta-learning idea, and it is indispensable to AAI's Criterion~1.
An AAI system that does not, in some form, meta-learn is a system
whose self-configuration has no mechanism.

\paragraph{Where AAI extends meta-learning.}
Meta-learning, as typically formulated, presupposes a task
distribution that is fixed, available in advance, and
representative of the deployment distribution. The meta-learner's
generalization is generalization \emph{within} the meta-training
distribution; out-of-meta-distribution performance is no more
guaranteed than out-of-distribution performance in ordinary
learning. AAI, by contrast, requires behavior under regime shift
(Criterion~2), which may include shifts outside the task
distribution on which meta-learning was performed. The mechanism
for handling such shifts is not internal to standard meta-learning
formulations, and its construction is one of AAI's central
engineering problems.

Meta-learning also, in its typical formulation, adapts
\emph{weights} or \emph{representations} while holding
\emph{architecture} and \emph{optimizer structure} fixed. AAI's
self-configuration extends to these structural choices when doing
so improves performance on the task. The meta-learning literature
has begun to address structural adaptation, but the integration is
not yet standard, and AAI treats it as required rather than
optional.

\subsection{AutoML}

AutoML names the body of work that automates the construction and
tuning of machine-learning pipelines: feature engineering, model
selection, hyperparameter optimization, ensemble construction, and
the scheduling of these operations. Its practical successes are
substantial. Commercial AutoML systems now deliver, on well-defined
tabular tasks, performance that matches or exceeds that of
experienced human practitioners, at a fraction of the human cost.

\paragraph{What AAI takes from AutoML.}
The commitment to removing humans from configuration decisions is
shared. AutoML has produced the practical infrastructure --- search
spaces, surrogate models, multi-fidelity evaluation, warm-starting
strategies --- that any AAI system will, in part, need. The
empirical finding that automated pipelines can outperform tuned
human ones on defined tasks is direct evidence for the feasibility
of shifting configuration decisions into the system.

\paragraph{Where AAI extends AutoML.}
The extension is of two kinds. First, AutoML typically operates in
\emph{construction time}: a pipeline is assembled and tuned, then
deployed as a fixed artifact. The tuning process, while automated,
is not itself part of the deployed system's adaptive behavior, and
when the deployment distribution shifts, re-invocation of AutoML
is a human-scheduled event. AAI requires the tuning loop to be
internalized, so that re-configuration occurs in response to
regime detection without external scheduling.

Second, and more deeply, AutoML's search procedures are themselves
configurations --- choice of search algorithm, choice of search
space, choice of surrogate, choice of budget --- that are set by
humans and not adapted to the task. By Principle~\ref{prin:conservation-tuning},
the tuning cost is relocated, not eliminated. AAI requires that the
system's choice of how to search its configuration space be itself
conditioned on the task, which is the recursion AutoML has largely
declined to take.

\subsection{Continual Learning}

Continual learning (also called lifelong learning) addresses the
problem of acquiring knowledge over a sequence of tasks without
destructive interference with earlier tasks. Its central
phenomenon is catastrophic forgetting, and its central techniques
are regularization-based (penalizing changes to parameters
important for earlier tasks), replay-based (interleaving old
examples with new), and architecture-based (allocating new
capacity to new tasks while preserving old).

\paragraph{What AAI takes from continual learning.}
Criterion~4 (preservation under revision) is, essentially, the
continual-learning desideratum. An AAI system that adapts to a new
regime at the cost of its competence on earlier regimes fails a
test that continual-learning research has made central. The
techniques developed in this program --- particularly the
architecture-based ones, which allocate separate capacity to
distinct regimes --- are directly usable as components of AAI.

\paragraph{Where AAI extends continual learning.}
Continual learning typically assumes that task boundaries are
given, either as explicit task identifiers or as a clean schedule
of task arrivals. When task boundaries are latent --- as they are
in most deployment environments --- standard continual-learning
methods either degrade or require the addition of a regime-detection
mechanism that is not part of their formulation. AAI requires
regime detection (Criterion~2) \emph{and} preservation under
revision (Criterion~4) simultaneously, and integrates them: the
detected regime informs which adaptations are applied, and the
preservation mechanism ensures that adaptations to one regime do
not overwrite another.

Continual learning also typically holds the learner's
\emph{architecture} and \emph{objective} fixed across tasks,
adapting only parameters. AAI's self-configuration permits
adaptation of these structural elements when the task warrants,
subject to the same preservation constraint.

\subsection{Neural Architecture Search}

Neural architecture search (NAS) automates the design of model
architectures, treating architecture as an optimization variable.
Its progress has been considerable, with searched architectures
now routinely competitive with hand-designed ones on benchmark
tasks, and with search procedures whose cost has fallen by orders
of magnitude relative to early work.

\paragraph{What AAI takes from NAS.}
The premise that architecture is a decision variable rather than a
fixed choice is shared. The search spaces, the differentiable
relaxations, the weight-sharing strategies, and the
performance-prediction models developed in NAS are usable
components of AAI's self-configuration machinery. Without NAS-like
mechanisms, the idea of task-conditioned architectural adaptation
remains abstract.

\paragraph{Where AAI extends NAS.}
Like AutoML, NAS typically operates in construction time, with
search and deployment temporally separated. AAI internalizes the
search, requiring that architectural choices be revisable at
deployment time when regime shifts warrant. Like continual
learning's, NAS's standard formulation assumes a fixed task;
multi-task and task-conditioned NAS exist but are not the default,
and they are not, in the general case, integrated with regime
detection or preservation-under-revision. AAI requires the
integration.

\subsection{Foundation Models as a Special Case}

A word is owed, before concluding this section, about foundation
models. Large pretrained models exposed through a prompting
interface are sometimes described as a realization of many of the
capabilities enumerated above: they are ``meta-learners'' in the
sense that in-context demonstrations shape their behavior; they are
``architecturally flexible'' in the sense that a single model serves
many tasks; they are ``continual'' in the sense that a fine-tuning
pass updates them without full retraining.

This description is partly accurate and largely misleading. The
in-context adaptation of a foundation model is a form of
task-conditioned configuration, and counts toward AAI's Criterion~1
to the extent that the conditioning is performed by the model
rather than by a human prompt-engineer. But the other three
criteria are poorly satisfied by default. Regime detection
(Criterion~2) is not part of the standard foundation-model stack;
regime-aware behavior is imposed, where it exists, by the pipeline
around the model, not by the model itself. Structural minimality
(Criterion~3) is failed comprehensively, since the number of
human-set hyperparameters in a production foundation-model stack
(from pretraining corpus composition through prompt template to
decoding strategy) is very large. Preservation under revision
(Criterion~4) is frequently violated: fine-tuning on new tasks
degrades performance on prior ones, and the mitigations remain
partial.

Foundation models are therefore best understood as a particularly
capable class of ANI systems, most of whose adaptive behavior is
performed by the human-operated pipeline surrounding them rather
than by the models themselves. This is not a dismissal; foundation
models are the most capable components we have, and any AAI system
built in the next decade will almost certainly include one. The
point is that including a foundation model is not the same thing
as being an AAI system, and the distinction matters for what the
remaining sections of this book set out to build.

\section{Evaluation Axes for Adaptivity}
\label{sec:evaluation-axes}

The criteria of Section~\ref{sec:positioning-aai} specify what AAI
\emph{is}; they do not yet specify how adaptivity is \emph{measured}.
Benchmarking practices developed for ANI do not transfer cleanly.
A benchmark that reports accuracy on a fixed test set rewards
systems that were carefully tuned for that test set, whether the
tuning was performed by a human or by the system itself; the
benchmark cannot tell the difference, and so cannot distinguish
ANI from AAI on its own terms. A useful evaluation methodology for
adaptive systems must therefore measure not only performance but
the \emph{process by which performance is achieved}, and must
penalize systems that purchase performance through configuration
cost the benchmark fails to charge.

This section introduces four axes along which adaptivity is
evaluated. Each axis corresponds to a dimension of
Definition~\ref{def:aai-operational} and to one of the four
distinguishing criteria of Section~\ref{sec:positioning-aai}. A full
evaluation of a candidate AAI system reports a position on each axis,
and the position is, in each case, a relation between performance
and configuration cost rather than a single number.

\subsection{Axis 1: Parameter Minimality}

\begin{definition}[Parameter minimality]
\label{def:parameter-minimality}
The \emph{parameter minimality} of a system $S$ on a task
distribution $\mathcal{T}$ is the smallest set
$\mathcal{H}^{\ast}(S, \mathcal{T}) \subseteq \mathcal{H}(S)$ of
hyperparameters such that, when the values of hyperparameters
in $\mathcal{H}^{\ast}$ are held fixed and those outside are
permitted to be set by $S$, performance on $\mathcal{T}$ remains
within a stated tolerance of the system's best achievable
performance. Parameter minimality is measured by the cardinality
of $\mathcal{H}^{\ast}$, weighted by the operational significance
of its elements.
\end{definition}

\noindent
The intuition is that a more adaptive system \emph{needs less from
its user}: fewer decisions are required before it can be deployed,
fewer are required to keep it deployed, and fewer are required to
move it to a new task. The axis inverts the natural instinct of
benchmark design, which tends to report performance at optimum
tuning and to treat the tuning effort as invisible. Parameter
minimality measures what the system has absorbed by measuring
what it no longer asks.

\paragraph{Reporting requirements.}
Honest reporting on this axis requires the full hyperparameter set
$\mathcal{H}(S)$ to be enumerated, including the hyperparameters of
any sub-systems (optimizers, tokenizers, retrieval components, and
so on) whose values might otherwise be inherited from defaults
without notice. Defaults are human-chosen values and count as
human decisions unless the system demonstrably re-selects them
from first principles at deployment time. A system whose
``automatic'' behavior depends on a particular weight-decay setting
chosen by a developer in 2017 is not adaptive with respect to that
hyperparameter, however invisible its presence may have become.

\paragraph{Relation to capacity.}
Parameter minimality is not capacity minimization. A large model
with few exposed hyperparameters scores better on this axis than a
small model with many. The axis measures the interface between
system and user, not the internal scale of the system. This
distinction matters because it resists the conflation of
``simplicity'' with ``smallness''; an adaptive system can be large.

\subsection{Axis 2: Transferability}

\begin{definition}[Transferability]
\label{def:transferability}
The \emph{transferability} of a system $S$ from a source task
distribution $\mathcal{T}_0$ to a target distribution $\mathcal{T}_1$
is the performance of $S$ on $\mathcal{T}_1$ as a function of the
adaptation budget permitted between observation of $\mathcal{T}_1$
and evaluation. The adaptation budget is measured in both
\emph{data} (examples from $\mathcal{T}_1$) and \emph{intervention}
(human decisions). A system with higher transferability achieves
higher target performance at lower adaptation cost.
\end{definition}

\noindent
This axis captures what the task-transfer family of AGI definitions
was reaching for (Section~\ref{sec:agi-premature}) and makes it
quantitative. It is not enough to report a transfer-learning
accuracy number; the axis requires that the accuracy be paired with
the adaptation budget that produced it, with budgets measured in
comparable units across systems.

\paragraph{Two-dimensional reporting.}
Transferability is, in general, a curve in a two-dimensional space
whose axes are data budget and intervention budget. A system that
requires many examples but no human decisions is better, on this
axis, than one that requires few examples but many human decisions,
because the former absorbs the configuration burden the latter
externalizes. Reporting single-number summaries (e.g., few-shot
accuracy) hides the intervention axis and flatters systems that
rely on it.

\paragraph{Source-to-target distance.}
A full evaluation reports transferability across a range of
source-to-target distances, from near transfer (small
distributional shift) to far transfer (major distributional shift).
Systems that transfer well near and poorly far exhibit a different
kind of adaptivity from systems whose performance degrades
smoothly with distance, and the distinction is obscured by any
evaluation that fixes the distance.

\subsection{Axis 3: Regime Awareness}

\begin{definition}[Regime awareness]
\label{def:regime-awareness}
The \emph{regime awareness} of a system $S$ is measured by three
quantities: (a) its \emph{detection rate} for regime shifts in its
operating environment; (b) its \emph{response latency} between
detection and behavioral change; and (c) its \emph{false-positive
rate} of purported detections that do not correspond to genuine
shifts. Higher regime awareness means detecting more real shifts,
responding faster, and hallucinating fewer.
\end{definition}

\noindent
This axis operationalizes Criterion~2. Regime awareness is not a
property that emerges for free from in-distribution training; it
requires explicit machinery whose evaluation has no counterpart in
standard benchmarks. A benchmark test set drawn from the same
distribution as training provides no information about whether the
system would have noticed, had the distribution changed.

\paragraph{Evaluation protocol.}
Regime awareness is evaluated by exposing the system to a stream
that includes both within-regime and cross-regime segments,
without informing the system which is which, and measuring the
three quantities above. The stream must include both clean
transitions (sharp shifts) and ambiguous ones (gradual drift),
because systems that handle one class of transition do not
necessarily handle the other.

\paragraph{Interaction with performance.}
High regime awareness is easy to achieve if one is willing to
accept high false-positive rates; such systems re-configure
constantly and pay for it in stability. The axis therefore demands
joint reporting with the other axes, and in particular with the
preservation axis below.

\subsection{Axis 4: Self-Reconfiguration}

\begin{definition}[Self-reconfiguration]
\label{def:self-reconfiguration}
The \emph{self-reconfiguration} of a system $S$ is the extent to
which, in response to detected regime shift or task change, $S$
modifies its own configuration --- including architectural
structure, optimizer, objective, and inference strategy --- and
does so without external scheduling, without destroying prior
competence (Criterion~4), and within a bounded time and resource
budget.
\end{definition}

\noindent
Self-reconfiguration is where adaptation becomes \emph{action}:
detection without reconfiguration is diagnosis without treatment,
and reconfiguration without detection is drift. The axis measures
the reconfiguration mechanism's effectiveness, its scope (which
components of the system it can reconfigure), its cost (the
compute and data required per reconfiguration event), and its
preservation properties (the extent to which it degrades
performance on prior regimes).

\paragraph{Scope of reconfiguration.}
Systems differ substantially in what they can reconfigure. A system
that can only adjust parameters is less adaptive, on this axis,
than one that can adjust optimizer settings; less adaptive still
than one that can adjust architectural structure. Scope is
measured by enumeration against a stated reference set of
reconfigurable elements, and reporting requires the enumeration to
be explicit rather than implicit.

\paragraph{Cost per reconfiguration.}
A system that reconfigures cheaply can afford to reconfigure often;
one that reconfigures expensively must be conservative about
doing so. The cost per reconfiguration therefore interacts with
regime awareness: high awareness plus high cost produces either
instability or paralysis, and the design of the reconfiguration
mechanism must be evaluated against the detection mechanism it is
paired with.

\subsection{Joint Reporting and the Adaptivity Profile}

None of the four axes is sufficient on its own. A system can
exhibit excellent parameter minimality by exposing few
hyperparameters while performing poorly; excellent transferability
on near tasks while failing on far ones; excellent regime
awareness coupled with destructive self-reconfiguration; excellent
self-reconfiguration with no way of knowing when to trigger it. A
useful evaluation of a candidate AAI system reports a position on
each axis and treats the joint position as the system's
\emph{adaptivity profile}.

\begin{definition}[Adaptivity profile]
\label{def:adaptivity-profile}
The \emph{adaptivity profile} of a system $S$ with respect to a
task distribution $\mathcal{T}$ and context space $\mathcal{C}$ is
the tuple
\[
\Pi(S; \mathcal{T}, \mathcal{C}) \;=\; \bigl(\mathcal{M}(S), \mathcal{X}(S), \mathcal{R}(S), \mathcal{F}(S), P(S)\bigr),
\]
where $\mathcal{M}$ denotes parameter minimality,
$\mathcal{X}$ transferability, $\mathcal{R}$ regime awareness,
$\mathcal{F}$ self-reconfiguration, and $P$ the system's
task-performance level under these conditions.
\end{definition}

\noindent
Two systems are comparable on adaptivity to the extent that their
profiles can be dominance-ordered: system $S_1$ is more adaptive
than $S_2$ if $\Pi(S_1)$ Pareto-dominates $\Pi(S_2)$ on all five
components. When profiles do not dominance-order --- which will be
the common case --- the comparison is explicit about which axes
favor which system, rather than collapsing the disagreement into a
single contested number.

This evaluation framework will be taken up in full in
next chapters, where the protocols are specified in
detail and applied to representative systems. For the remainder of
the present chapter, the framework's role is conceptual: it fixes
what the rest of the book is trying to measure, and so fixes the
criterion by which later proposals --- architectural, algorithmic,
evaluative --- will be judged. A proposal that improves performance
without improving the adaptivity profile is not, by the standards of
this book, a proposal for AAI. It is a proposal for a more capable
ANI, and the distinction, though subtle in the literature, is
sharp in the framework developed here.

\chapter{The Principle of Parametric Minimality}
\label{chap:parametric-minimality}

\epigraph{%
\emph{Perfection is achieved, not when there is nothing more to add,
but when there is nothing left to take away.}%
}{Antoine de Saint-Exup\'ery, \emph{Terre des hommes}, 1939}

\noindent
I previously argued that adaptivity is measured not
by what a system does but by what it no longer asks its user to
decide. That formulation is, so far, rhetorical. To give it operational
content, the quantity ``decisions asked of the user'' must be made
precise, its relation to system capability must be characterized, and
its minimization must be stated as a design objective against which
proposals can be tested.

This chapter undertakes the first and second of those tasks. Section
\ref{sec:param-count-proxy} argues that the combined count of
hyperparameters and human-set architectural choices, appropriately
weighted, is a usable proxy for a system's dependence on human
configuration effort --- not an ideal measure, but the best one
currently available, and one whose deficiencies are themselves
instructive. Section~\ref{sec:minimal-tuning-objective} formalizes
``minimal tuning'' as an objective, gives it a precise statement,
analyzes its properties, and examines the trade-offs it induces
against competing objectives such as performance and reliability.
The remaining sections of the chapter (treated in subsequent parts
of this book) develop the algorithmic consequences of adopting this
objective as a design principle.

The principle this chapter states can be expressed in a single
sentence: \emph{a system's dependence on human configuration effort
is a cost, and that cost should be driven toward its irreducible
minimum subject to the system's performance and reliability
constraints}. The sentence is short; its consequences are not.

\section{Why Parameter Count Is a Proxy for Human Dependence}
\label{sec:param-count-proxy}

\subsection{The Question Behind the Question}

The statement that adaptivity should be measured by reductions in
human configuration effort raises an immediate methodological
question: how is human configuration effort to be measured? The
question is not trivial. Effort is a psychological and organizational
quantity; it depends on the practitioner, the institution, the
available tooling, and the accumulated tacit knowledge of the
community. A hyperparameter that a seasoned practitioner sets
effortlessly may consume weeks of a beginner's attention. A
configuration decision that is invisible in one institution (because
a standard value has been adopted) is a fraught choice in another.
Any measure that tried to capture effort directly would be so
contingent on context as to be useless for comparing systems.

The alternative is to measure not effort but \emph{surface area}:
the count and character of decisions the system exposes to human
configuration, regardless of how difficult those decisions are in
any particular setting. Surface area has three advantages over
effort. It is observable: one can inspect a system and enumerate
the decisions it requires. It is comparable: the enumeration
produces a number (or a structured quantity) that can be set
alongside the corresponding number for another system. And it is
manipulable: a design change that removes a decision from the
system's interface reduces its surface area, whether or not the
decision was difficult in the first place.

The proposal of this section is that parameter count --- understood
broadly to include both hyperparameters in the conventional sense
and architectural choices that function as configuration decisions
--- serves as a usable surface-area measure. The proposal is not
that parameter count \emph{is} human dependence; the two are
conceptually distinct. It is that parameter count \emph{proxies}
human dependence, in the sense that reductions in the former
systematically produce reductions in the latter, and that systems
which expose fewer decisions to their users systematically demand
less of them. The proxy is imperfect and its imperfections are
discussed below, but no better proxy is available, and the
imperfections are of a kind that conservative design can work
around.

\subsection{Hyperparameters as Latent Human Decisions}

A hyperparameter, in the sense used here, is any numerical or
categorical value that governs a system's behavior and is set
before the system processes any task-relevant input. The
learning rate of a gradient optimizer is a hyperparameter. The
regularization coefficient of a penalized regression is a
hyperparameter. The number of nearest neighbors in a $k$-NN
classifier, the temperature of a softmax, the depth of a search
tree, the beam width of a decoding procedure, the dropout rate,
the batch size, the choice of activation function, the random seed
--- all are hyperparameters by this definition, and all share the
property that a human, somewhere in the system's history, chose
their values.

The choice may be explicit, in the sense that a developer wrote
the value into a configuration file with deliberation, or
implicit, in the sense that a default was inherited from a library
or a predecessor system. The distinction between explicit and
implicit choice is operationally important but conceptually
secondary: an implicit default is still a choice, made by someone
at some point, and its presence in the system is evidence that a
decision was made and that its revision would require decision-making.
A system with fifty hyperparameters, of which forty-eight are set
to their library defaults, is a system with fifty human-set values,
not two.

This point deserves emphasis because it is commonly evaded. Papers
that describe their hyperparameter choices with phrases like ``we
used standard values'' or ``default settings were employed''
understate the surface area of their systems. The standard value
was chosen, by someone, under conditions that may or may not hold
for the new deployment; its invisibility in the paper does not
reduce the system's dependence on it. An honest accounting requires
all hyperparameters to be enumerated, including the ones whose
values are inherited, and their values to be reported whether or
not they differ from defaults. The adaptivity index presupposes such an accounting, and
its reliability depends on the accounting being complete.

\begin{principle}[The full hyperparameter census]
\label{prin:full-census}
The hyperparameter set of a system, for the purpose of measuring
human dependence, consists of every value that governs the system's
behavior and is not computed by the system from task-relevant
input at use time --- regardless of whether the value was set by
explicit deliberation, inherited from a library default, copied
from a predecessor system, or assumed without discussion. Defaults
are human decisions by other authors; they do not become the
system's own decisions by virtue of their invisibility.
\end{principle}

\subsection{Architectural Choices as Discrete Hyperparameters}

Hyperparameters, narrowly construed, are continuous or
low-cardinality categorical values sitting in otherwise-fixed
structural contexts. But a great many human decisions concerning
machine-learning systems are architectural: the number of layers,
the width of each layer, the connectivity pattern between layers,
the attention mechanism's configuration, the presence or absence
of residual connections, the tokenizer's vocabulary size, the
retrieval system's index structure. These decisions are made by
humans, govern system behavior, and are fixed before the system
processes any input. They are therefore hyperparameters in the
sense of the previous subsection, and the surface-area accounting
should include them.

The convention in the literature of treating architecture as
``given'' and hyperparameters as ``tuned'' is an artifact of
separate research communities (designers of architectures,
optimizers of architectures' scalar settings) rather than a
principled distinction. From the standpoint of the user, a system
whose architecture was chosen by its developers and whose scalar
hyperparameters are chosen by the user is a system that depends on
both sets of choices, and the user has no principled ground on
which to treat one set as more fundamental than the other. An
adaptivity measure that counts scalar hyperparameters and ignores
architectural ones is measuring the surface area of a selected
face of the system rather than the surface area of the system.

\begin{definition}[Generalized hyperparameter set]
\label{def:generalized-hyp}
The \emph{generalized hyperparameter set} $\mathcal{H}(S)$ of a
system $S$ is the set of all decisions --- numerical, categorical,
or structural --- whose values govern $S$'s behavior and which are
fixed before $S$ processes any task-relevant input. Architectural
choices, connectivity patterns, vocabulary definitions,
preprocessing pipelines, and default values of nominally
``internal'' library parameters are all elements of $\mathcal{H}(S)$,
on the same footing as scalar hyperparameters in the conventional
sense.
\end{definition}

\subsection{The Proxy Relationship}

Given a generalized hyperparameter set, the claim that its
cardinality proxies human dependence requires argument. Three
reasons support the claim; each has limitations that are treated
in the next subsection.

\paragraph{Reason 1: Decisions require deciders.}
Each element of $\mathcal{H}(S)$ must have been set to some value,
and the setting is an event in the world that required an agent to
perform it. In the overwhelming majority of contemporary systems,
that agent is a human. A system with $|\mathcal{H}(S)| = n$ has
therefore required at least $n$ human decisions in the course of
its preparation, whether those decisions were taken by the system's
developers, by the authors of inherited libraries, or by the user
at deployment time. Reducing $|\mathcal{H}(S)|$ reduces the number
of decisions that must be taken, and so reduces the aggregate
human labor embedded in the system's configuration.

\paragraph{Reason 2: Decisions propagate to users.}
Hyperparameters that are set centrally, once and for all, by the
system's developers impose less burden on users than hyperparameters
that users are expected to set themselves. But the distinction is
not as sharp as it appears. Centrally-set defaults are optimal for
some distribution of use cases and suboptimal for others; when the
user's distribution differs from the developer's, the user is in
the position of having to decide whether to override the default
and, if so, to what value. This decision is itself configuration
effort, and it scales with the count of defaults whose suitability
the user must assess. Even where the user does not revise a
default, the act of inspecting the default and accepting it is
decision-making. Surface area therefore propagates from developer
to user; it is not extinguished by central setting.

\paragraph{Reason 3: Decisions compound across deployments.}
A system deployed to a single context can, in principle, have its
hyperparameters tuned once and then left alone. A system deployed
across many contexts requires, at minimum, the question ``do the
tuned values still apply?'' to be asked for each context. If
$|\mathcal{H}(S)|$ is large and deployments are many, the aggregate
decision cost is $|\mathcal{H}(S)| \times (\text{deployments})$, and
reducing the first factor reduces the cost at every deployment.
A system whose hyperparameter count is small can be deployed widely
at low marginal cost; a system whose count is large cannot,
whatever the competence of the individual deployment.

These three reasons together justify treating $|\mathcal{H}(S)|$,
suitably weighted, as a proxy for the human dependence of $S$. The
proxy is not perfect, and the imperfections are important enough
to treat directly.

\subsection{Imperfections of the Proxy}

Three classes of imperfection limit the proxy's accuracy.

\paragraph{Not all hyperparameters are equally burdensome.}
A hyperparameter whose optimal value is robust across tasks
imposes less burden than one whose optimal value is task-dependent,
because the former can be set once and forgotten while the latter
must be revisited. Similarly, a hyperparameter whose setting
affects a small fraction of system behavior imposes less burden
than one whose setting affects a large fraction. An unweighted count
is therefore a coarse measure; a weighted count, with weights
reflecting task-sensitivity and operational significance, is a
finer one. The adaptivity index of
Definition~\ref{def:adaptivity-index} incorporates such weights,
and the parameter-minimality axis of
Definition~\ref{def:parameter-minimality} presupposes their use.

The weights themselves, however, must come from somewhere, and
choosing them is a meta-decision of the kind
Principle~\ref{prin:conservation-tuning} warns about. There is no
principled way to eliminate this meta-decision entirely; what can
be done is to make the weights explicit, to evaluate systems under
multiple weight schemes to test robustness, and to prefer weight
schemes with intersubjective support (consensus among
practitioners, or derivation from measured task-sensitivity
studies) over idiosyncratic ones.

\paragraph{Hidden surface area.}
Some decisions that affect system behavior do not appear in any
conventional hyperparameter listing because they are made at a
layer the practitioner does not ordinarily inspect: the choice of
numerical precision, the order of floating-point operations, the
scheduler of the GPU kernels, the compilation flags of the
underlying libraries. These decisions can, in pathological cases,
have outsized effects on system behavior, and a surface-area
measure that ignores them understates dependence. An honest
accounting must, at minimum, note that such decisions exist and
that their values are held constant across the comparisons in
which the measure is used. The surface-area measure is most
reliable when it is applied to systems whose substrates are
comparable, and becomes less reliable as substrate differences
grow.

\paragraph{Substitution effects.}
Reducing $|\mathcal{H}(S)|$ by making a hyperparameter
task-conditioned (so that it is computed from task-relevant input
rather than set in advance) can introduce new hyperparameters
governing the conditioning mechanism itself. If the new count
exceeds the old count, the surface-area reduction was illusory.
Principle~\ref{prin:conservation-tuning} names this phenomenon;
the proxy is vulnerable to it whenever the conditioning mechanism
is not itself subject to the same accounting. Honest reporting
requires the hyperparameters of the adaptation mechanism to be
counted against the system's surface area, with the test of
progress being whether the total count decreases after the
conditioning is introduced. In practice, the test is often passed
--- well-designed conditioning mechanisms replace many
hyperparameters with few --- but the test must be applied, and
cannot be assumed.

\subsection{Against the Inevitability of High Surface Area}

A frequent objection to the surface-area proxy is that high
hyperparameter counts are inevitable for capable systems. The
objection runs: powerful systems are complex, complex systems
have many configurable elements, therefore capable systems have
high surface areas, and any proxy that penalizes high surface
areas penalizes capability.

The objection has the structure of a plausible argument and the
status of an assumption. It would be true if hyperparameters were
a necessary consequence of internal complexity; it is false
because internal complexity can, in well-designed systems, be
\emph{hidden} from the user rather than exposed to the user. The
human nervous system is vastly more complex than any deployed AI
system, and exposes a configuration interface of substantially
lower cardinality; evolution, as a designer, had strong incentives
to absorb configuration into the system rather than exposing it to
any external tuner. The contrast suggests that the coupling between
internal complexity and interface surface area is a property of how
systems are built, not a law of nature, and that design choices
which reduce the coupling are possible in principle.

The empirical record supports this suggestion. Compiler technology
has, over decades, substantially reduced the number of decisions
programmers must make about code generation, at the cost of
substantial internal complexity in the compiler. Modern database
systems expose configuration interfaces far smaller than their
predecessors' while offering greater capability, through
internalization of decisions that were once the user's burden.
Operating-system schedulers have reduced the user-facing parameters
of process management from dozens to effectively none, while
becoming internally more sophisticated. In each case, the
engineering achievement was to absorb decisions into the system,
and in each case, the achievement produced systems whose
capability grew while whose human dependence fell.

No equivalent maturation has occurred for machine-learning systems,
and the objection above is often, on inspection, a restatement of
the field's current practice rather than an argument for its
inevitability. The surface-area proxy does not penalize capability;
it penalizes the exposure of internal complexity to the user, and
rewards designs that keep internal complexity internal. That this
is a more demanding standard than the field has hitherto applied
is the point.

\section{Formalizing Minimal Tuning as an Objective}
\label{sec:minimal-tuning-objective}

\subsection{From Proxy to Principle}

If surface area is a usable proxy for human dependence, and if
reductions in human dependence are what adaptivity requires, then
the design of adaptive systems should take surface-area reduction
as an explicit objective. This subsection formalizes that objective,
analyzes its mathematical structure, and states the principle that
governs its application.

The naive formulation --- ``minimize $|\mathcal{H}(S)|$'' --- is
useless on its own because the minimum is zero, achieved by the
trivial system that does nothing. Minimality must be constrained by
performance: the system must still accomplish its tasks. It must
also be constrained by reliability: the reduction in surface area
must not come at the cost of the system's ability to perform
predictably across contexts. And it must be constrained by honest
accounting: reductions that merely relocate hyperparameters into
mechanisms not counted by the accounting are not genuine reductions.

\begin{definition}[Minimal-tuning objective]
\label{def:min-tuning}
Let $\mathcal{T}$ be a task distribution, $\mathcal{C}$ a context
space, and $P(S; \mathcal{T}, \mathcal{C})$ a performance measure
of system $S$ on $\mathcal{T}$ under contexts in $\mathcal{C}$. The
\emph{minimal-tuning objective} for the design of an adaptive
system over $(\mathcal{T}, \mathcal{C})$ is
\[
S^{\ast} \;=\; \arg\min_{S \in \mathcal{S}} \; \Phi(\mathcal{H}(S))
\quad \text{subject to} \quad P(S; \mathcal{T}, \mathcal{C}) \geq P_{\min},
\]
where $\mathcal{S}$ is a class of admissible systems,
$\Phi(\mathcal{H}(S)) = \sum_{h \in \mathcal{H}(S)} w(h)$ is a
weighted surface-area measure, $w(h) \geq 0$ weights element $h$
by its task-sensitivity and operational significance, and
$P_{\min}$ is a required performance threshold.
\end{definition}

\noindent
The objective admits several observations. It is a
\emph{constrained minimization}, not a weighted sum; the
performance requirement is a constraint rather than a term traded
off against surface area. This structure reflects the practical
reality that performance below a threshold is unacceptable
regardless of how small the surface area becomes, while performance
above the threshold offers diminishing returns relative to further
surface-area reductions. Systems whose performance varies
continuously with surface area can be analyzed by varying
$P_{\min}$ and tracing the Pareto frontier; the constrained form
is the analytical primitive from which such analyses are built.

The weights $w(h)$ are required because, as the previous section
noted, not all hyperparameters are equally burdensome. In the
simplest case, $w(h) = 1$ for all $h$ and $\Phi$ reduces to the
cardinality $|\mathcal{H}(S)|$. More informative weights are
available when task-sensitivity measurements have been performed:
$w(h)$ may be set proportional to the measured variance of
performance with respect to $h$, or to the measured frequency with
which $h$ is revised across deployments, or to a combination of
these.

The admissible class $\mathcal{S}$ is a nontrivial modeling choice.
In strict applications, $\mathcal{S}$ is constrained to systems
whose implementation admits honest hyperparameter accounting;
systems whose internal decisions cannot be enumerated are excluded,
on the ground that their surface area cannot be measured. In looser
applications, $\mathcal{S}$ may include any system that performs
the task, and accounting is performed post hoc at whatever
resolution the system's implementation permits. The strict form is
preferable for research purposes, the loose form for comparison of
deployed systems against baselines.

\subsection{Recursion and the Meta-Hyperparameter Problem}

Definition~\ref{def:min-tuning} is not, on its own, enough. The
objective defines what an adaptive system should minimize, but a
system designed by solving the objective is a system whose design
was itself configured by some procedure, and that procedure had
hyperparameters of its own. If the procedure is an AutoML-style
search, it had choices of search algorithm, search space, surrogate
model, and budget. If the procedure is a meta-learner, it had
choices of meta-training task distribution, meta-optimization
algorithm, and inner-loop structure. If the procedure was carried
out by a human team, the team made numerous decisions whose values
are embedded in the resulting system.

Principle~\ref{prin:conservation-tuning} (the conservation of tuning
burden) implies that the design procedure's surface area must be
counted against the system it produces, or else the formalization
of Definition~\ref{def:min-tuning} is vulnerable to arbitrary
relocation of hyperparameters out of the system and into the
procedure that built it. The honest form of the minimal-tuning
objective therefore extends to the procedure as well.

\begin{definition}[Honest minimal-tuning objective]
\label{def:honest-min-tuning}
Let $S$ be a system and $D$ be the design procedure that produced
$S$, with generalized hyperparameter sets $\mathcal{H}(S)$ and
$\mathcal{H}(D)$ respectively. The \emph{honest minimal-tuning
objective} is the constrained minimization
\[
(S^{\ast}, D^{\ast}) \;=\; \arg\min_{(S, D) \in \mathcal{S} \times \mathcal{D}}
\; \Phi(\mathcal{H}(S)) + \lambda\, \Phi(\mathcal{H}(D))
\]
subject to $P(S; \mathcal{T}, \mathcal{C}) \geq P_{\min}$ and to the
constraint that $D$ produces $S$. The constant $\lambda \geq 0$
weights the design procedure's surface area relative to the
system's.
\end{definition}

\noindent
The weighting by $\lambda$ reflects a substantive design decision.
Setting $\lambda = 1$ treats surface area at the design stage and
surface area at the deployment stage as equally costly. Setting
$\lambda = 0$ treats design-stage surface area as free, reproducing
the AutoML stance that the search procedure's own configuration is
outside the system's accounting. Setting $\lambda > 1$ treats
design-stage surface area as more costly than deployment-stage
surface area, on the ground that design-stage decisions propagate
across all subsequent deployments while deployment-stage decisions
are local to one context.

The right choice of $\lambda$ depends on the intended use of the
system. A system built for a single deployment should use
$\lambda$ close to zero: the design procedure runs once, its
configuration is amortized, and what matters is the interface the
user faces thereafter. A system built for general use, to be
deployed repeatedly across contexts, should use $\lambda$ close to
or above one: the design procedure's configuration is part of what
consumers of the system inherit, and its reduction is part of the
system's generality. The critical point is that the choice must be
stated: omitting the design procedure's surface area from the
accounting is equivalent to setting $\lambda = 0$, and this choice
should be made deliberately rather than by default.

Recursion does not, in general, terminate. The design procedure $D$
was itself designed, by a procedure $D'$, and so on. In practice,
the regression is cut off at a level where further accounting
provides diminishing returns --- typically at the level of design
choices taken by the research community over decades, which are
treated as shared background rather than as surface area attributable
to any particular system. This pragmatic cutoff is a necessary
concession, but its existence should be acknowledged. No system is
truly free of background configuration effort; the measure applies
to what is distinctive about the system under evaluation, holding
the community-wide background constant.

\subsection{Properties of the Objective}

The minimal-tuning objective has several properties worth noting,
each with consequences for how it can be used.

\paragraph{Monotonicity under absorption.}
If a design modification absorbs a hyperparameter $h \in \mathcal{H}(S)$
into a task-conditioned inference, without introducing new
hyperparameters that compensate, then $\Phi(\mathcal{H}(S))$
strictly decreases, and the modification is preferred by the
objective provided that performance is preserved. This property
gives the objective its teeth: it rewards precisely the design
moves that the surface-area-as-proxy argument calls for.

\paragraph{Non-monotonicity under structural change.}
Many design changes simultaneously add and remove hyperparameters
--- replacing one architectural module with another, say --- and
the effect on $\Phi$ depends on the net change in weighted count.
The objective is therefore not a greedy rule of ``always remove
hyperparameters''; it is a criterion for evaluating changes whose
structure may be complex. A change that adds three hyperparameters
while removing five is preferred, other things equal; a change that
adds five while removing three is not. The objective's value lies
in making such comparisons explicit rather than in dictating local
moves.

\paragraph{Non-identifiability.}
The objective does not uniquely identify a best system; many
systems may have the same $\Phi$ value and the same $P$ value.
Non-identifiability is not a defect of the objective but a
reflection of the underlying space: there are genuinely multiple
ways to build systems with comparable surface areas and comparable
performance, and the choice among them must be made on other
grounds (robustness, development cost, interpretability, alignment
with broader architectural commitments). The objective narrows the
design space; it does not collapse it to a point.

\paragraph{Context-dependence.}
The objective is defined with respect to a task distribution
$\mathcal{T}$, a context space $\mathcal{C}$, and a performance
threshold $P_{\min}$. A system that is minimal for one
$(\mathcal{T}, \mathcal{C}, P_{\min})$ tuple may not be minimal for
another. This context-dependence is important to acknowledge:
minimality is not absolute but relative to the stated scope of the
system's operation. A system that serves a narrow scope efficiently
should not be faulted for exposing hyperparameters needed only in
broader scopes, and a system that serves a broad scope should not
be rewarded for hiding hyperparameters under defaults that fail in
parts of its stated operating range.

\subsection{Trade-offs Against Performance and Reliability}

The constrained form of the objective asserts that performance is a
constraint, not a term to be traded away. But the structure of the
constrained problem still admits trade-offs, and understanding them
matters for applying the objective in practice.

\paragraph{The performance-minimality frontier.}
For any fixed design class $\mathcal{S}$, the minimum value of
$\Phi$ consistent with $P(S) \geq P_{\min}$ is a nondecreasing
function of $P_{\min}$. As the required performance rises, the set
of admissible systems shrinks, and the minimum achievable surface
area --- if it changes at all --- rises. The relation between
$P_{\min}$ and achievable $\Phi$ defines a \emph{frontier} in the
performance-minimality plane, and different deployment contexts
correspond to different points on this frontier.

An important empirical question, which the rest of this book takes
up in various forms, is the shape of this frontier. If it is steep
--- if small gains in required performance demand large increases
in surface area --- then the minimal-tuning objective is in severe
tension with capability, and its pursuit requires accepting
performance compromises that may not be tolerable in practice. If
it is shallow --- if large regions of the performance range are
achievable at nearly-minimal surface area --- then the objective
can be pursued with modest performance cost, and the principle
justifies strong claims. The shape of the frontier is, on the
evidence currently available, closer to shallow than to steep over
most of the practically-relevant performance range, but the
evidence is partial and the question remains open.

\paragraph{The reliability dimension.}
Performance is not the only criterion a deployed system must meet;
reliability --- the variance, tail behavior, and failure modes of
the system's performance across contexts --- is often equally
important. A system with small surface area and good mean
performance but occasional catastrophic failures is not preferable
to a system with larger surface area and uniformly adequate
performance. The minimal-tuning objective, as stated in
Definition~\ref{def:min-tuning}, does not capture this distinction,
and a full formulation must extend the performance constraint to
include reliability measures.

\begin{definition}[Reliability-constrained minimal tuning]
\label{def:reliable-min-tuning}
Let $R(S; \mathcal{T}, \mathcal{C})$ denote a reliability measure
of $S$ (such as worst-case performance, tail-conditional
performance, or variance across contexts). The
\emph{reliability-constrained minimal-tuning objective} is
\[
S^{\ast} \;=\; \arg\min_{S \in \mathcal{S}} \; \Phi(\mathcal{H}(S))
\]
subject to $P(S) \geq P_{\min}$ \emph{and} $R(S) \geq R_{\min}$,
where $R_{\min}$ is a required reliability threshold.
\end{definition}

\noindent
The addition of the reliability constraint changes the
optimization substantively. Hyperparameters whose presence
improves reliability (by permitting context-specific tuning that
prevents catastrophic failure in edge regimes) are, under the
reliability-constrained form, harder to absorb. Their absorption
must be accompanied by a demonstration that the absorbed mechanism
maintains reliability across the stated context space, not merely
that it matches mean performance. This raises the bar on what
counts as a legitimate surface-area reduction, and protects against
a pathology of naive minimality --- absorbing hyperparameters into
mechanisms that work on typical inputs and fail on atypical ones.

\subsection{The Principle, Stated}

The analysis of this chapter can be condensed into a principle that
guides the design moves of the chapters that follow.

\begin{principle}[Parametric minimality]
\label{prin:parametric-minimality}
An adaptive system's configuration surface area, measured by its
weighted generalized hyperparameter count and accounting honestly
for the surface area of the procedure that produced it, should be
minimized subject to performance and reliability constraints on
the system's stated operating range. Design modifications that
reduce the surface area are preferred, other things equal;
modifications that appear to reduce surface area but relocate it
into uncounted mechanisms do not count as reductions and should be
excluded from the accounting.
\end{principle}

\noindent
Three remarks complete the chapter.

First, the principle is prescriptive, not descriptive. It states
what adaptive systems ought to be designed to minimize, not what
existing systems in fact minimize. Most contemporary machine-learning
systems have not been designed under this principle; their high
surface areas are evidence of the principle's absence from their
design histories, not evidence against its correctness.

Second, the principle is comparative. It does not prescribe an
absolute target for surface area; it prescribes a \emph{direction}
of design effort and a \emph{criterion} for evaluating proposed
changes. A system whose surface area is smaller than its
predecessor's, holding performance and reliability, is an
improvement by the principle's standard; a system whose surface
area is larger is not, regardless of other merits.

Third, the principle does not stand alone. It is one of several
design principles for adaptive systems, and its application must be
balanced against the others --- particularly those governing regime
awareness, preservation under
revision, and compositional
structure. A system designed to
minimize surface area at the expense of the other principles may
be simpler to configure but less adaptive overall; the principles
act together, and their joint application is what produces AAI.

The next chapter develops the second of these principles, on regime
awareness: the system's capacity to recognize, without being told,
that its operating context has changed. Where parametric minimality
concerns what the system no longer asks of the user, regime
awareness concerns what the system notices on its own.

\section{Three Pathways to Minimality}
\label{sec:three-pathways}

\subsection{From Principle to Practice}

Principle~\ref{prin:parametric-minimality} prescribes a direction
for design effort but does not specify how that direction is to be
followed. The principle states that surface area should be
minimized; it does not say how the minimization is to be achieved.
The latter question --- by what mechanisms can a system absorb
into itself decisions that would otherwise fall to its user? ---
is the technical question that the bulk of this book addresses.
The answers, as the literature has developed them and as this book
will systematize them, fall into three broad pathways, each
characterized by a different locus of adaptation, a different
temporal relation to the system's deployment, and a different
class of design moves.

The pathways are not exclusive. A mature adaptive system will
typically combine elements of all three, and their interactions
are themselves a subject of analysis. But the pathways are
distinct enough that they can be treated separately for the
purpose of organizing the book, and distinct enough that
practitioners are well-served by having an explicit map of the
options available to them. This section provides that map.

\begin{definition}[Pathway to minimality]
\label{def:pathway}
A \emph{pathway to minimality} is a class of design mechanisms
that absorb generalized hyperparameters into the system by a
common organizing principle, identified by the locus at which the
absorption is effected and the temporal relation between the
absorption and the system's processing of input. The three
pathways treated in this book are: \emph{data- and task-aware
configuration} (locus: input-conditioned setting; temporality:
prior to or at the start of inference), \emph{structural and
evolutionary morphing} (locus: architecture-conditioned setting;
temporality: across episodes or developmental stages), and
\emph{in-training self-adaptation} (locus: trajectory-conditioned
setting; temporality: during the learning process itself).
\end{definition}

\noindent
The remainder of this section previews each pathway in turn,
indicating the part of the book that develops it in detail. The
treatment is necessarily brief; its purpose is to position the
detailed development in the context of the principle established
above and to make the relations among the pathways visible from
the start.

\subsection{Pathway I: Data- and Task-Aware Configuration}
\label{subsec:pathway-i}

The first pathway absorbs hyperparameters by computing their
values from features of the data or task at hand, rather than
fixing them in advance. The organizing observation is that many
configuration decisions which appear to require human judgment in
fact follow regularities that can be detected from inspection of
the data: the appropriate learning rate scales with the curvature
of the loss surface, the appropriate regularization strength
scales with the noise level of the labels, the appropriate model
capacity scales with the sample size and intrinsic dimensionality
of the input distribution, and so on. Where such regularities
exist and can be made explicit, the corresponding hyperparameters
need not be set by the user; they can be inferred by the system at
the start of processing.

Its mechanisms include: meta-features and
meta-learning frameworks that map dataset descriptors to
hyperparameter recommendations; task descriptors and conditioning
encoders that supply hyperparameter values as a function of task
embeddings; sample-size-aware capacity selection that grows model
capacity with available data; intrinsic-dimensionality estimation
that sets representation dimensionality from data geometry; and
noise-level estimation that sets regularization strength from
measured residual variance. Each mechanism replaces one or more
human decisions with a computation, and each can be evaluated
under Definition~\ref{def:honest-min-tuning} by checking whether
the computation's own configuration adds fewer hyperparameters
than the absorption removes.

The defining feature of this pathway is that the absorption is
effected \emph{before} the system commits to a configuration ---
the data, or a representation of the task, is examined and used
to set hyperparameters which then govern the rest of processing.
This temporal structure has two consequences. The first is that
the pathway is well-suited to settings where the data are
plentiful enough to support inference of the relevant
regularities, but the data may be heterogeneous across
deployments and the configuration cannot be fixed in advance.
The second is that, because the configuration is set once per
task or dataset and held thereafter, the pathway does not require
the system to re-adapt as it encounters new inputs within the same
task; this stability is desirable in many applications and
limiting in others.

The pathway's chief limitations are two. First, the regularities
on which it depends must exist and must be expressible in
forms the system can compute. Where the relation between data
features and optimal hyperparameter values is irregular,
non-stationary, or specific to combinations of features that the
meta-learner has not seen, the inferred values may be no better
than informed guesses --- and possibly worse than well-chosen
defaults. Second, the meta-learning machinery itself has
hyperparameters whose values must come from somewhere; the
honest accounting must include them, and the absorption is
genuine only when the count of meta-hyperparameters is smaller
than the count they replace. Both limitations are managed in
practice by careful empirical validation across diverse task
distributions, but neither is eliminated by clever design alone.

\subsection{Pathway II: Structural and Evolutionary Morphing}
\label{subsec:pathway-ii}

The second pathway absorbs hyperparameters by allowing the
system's structure --- its architecture, its connectivity, its
component composition --- to change in response to feedback from
its operation. The organizing observation is that many decisions
ordinarily fixed in advance (depth, width, branch structure,
choice of operator at each position) can be treated as variables
to be set by a search process that uses performance signals to
guide the search. Where the search is well-designed, decisions
that the user would otherwise have made are made instead by the
search, and the system's surface area is reduced by the
difference between the absorbed decisions and the search's own
configuration.

Its mechanisms span a range of techniques: neural
architecture search (NAS) in its various flavors (reinforcement
learning, evolutionary, gradient-based, weight-sharing); growing
networks that add capacity in response to under-fitting and prune
capacity in response to over-fitting; modular and composable
architectures whose component choice is determined by routing
mechanisms; and neuroevolutionary methods that treat architecture
as a population subject to selection pressure. Each mechanism
shares the property that the system's structure is not given but
constructed, and the construction is governed by signals from the
system's own performance rather than by external specification.

The defining feature of this pathway is that the locus of
adaptation is structural rather than parametric --- what changes
is what the system \emph{is}, not merely what it is set to. This
distinguishes the pathway sharply from Pathway~I, where the
structure is fixed and the absorbed hyperparameters are values
within it. Structural morphing absorbs decisions that Pathway~I
cannot reach: the depth of the network is not a value within a
network of fixed depth, and questions of connectivity are not
expressible as scalar settings. The temporal scale on which the
adaptation occurs is also longer; structural changes are typically
made across episodes or developmental stages rather than within a
single forward pass.

The pathway's chief limitations are three. First, the search
spaces involved are vast, and the search procedures require
computational budgets that may be unavailable in many contexts;
the surface-area savings at use time are paid for by computation
at design time. Second, the search procedures have their own
configuration --- search algorithm, search space definition,
budget, surrogate model, evaluation protocol --- and the honest
accounting requires this configuration to be counted against the
system's surface area, weighted by $\lambda$ in the sense of
Definition~\ref{def:honest-min-tuning}. Third, structural changes
can introduce non-stationarities that interact with parametric
learning in subtle ways, and the joint design of structural and
parametric adaptation is more delicate than either alone. 

\subsection{Pathway III: In-Training Self-Adaptation}
\label{subsec:pathway-iii}

The third pathway absorbs hyperparameters by allowing them to be
adjusted during the learning process itself, on the basis of
signals available from the trajectory of training. The organizing
observation is that many hyperparameters which are conventionally
fixed at the start of training (and which often need to be retuned
when learning fails) admit principled adjustment rules that use
training-time signals --- gradient norms, loss curvature, sample
difficulty, validation behavior --- to set their values
adaptively. Where such rules exist and are robust, the
hyperparameters they govern can be removed from the user-facing
interface entirely, replaced by the rule itself.

Its mechanisms include: adaptive optimizers whose effective
learning rates are set per-parameter from gradient statistics;
learning-rate schedules that respond to plateaus, validation
behavior, or curvature estimates; curriculum mechanisms that
order training examples by inferred difficulty; early-stopping
and patience rules that terminate training from validation
signals rather than fixed epoch counts; gradient clipping rules
whose thresholds adapt to observed gradient distributions; and
self-distillation and self-regularization mechanisms that use the
model's own predictions as targets for refinement. Each mechanism
treats a hyperparameter as a quantity to be tracked and updated
during learning rather than fixed at the start.

The defining feature of this pathway is that the temporal scale
of adaptation is finer than in either of the previous two: the
absorbing computation runs throughout training, not once before it
and not across coarse developmental stages. This finer scale gives
the pathway access to signals (intra-training trajectories) that
the other pathways cannot use, and supports adaptation of
hyperparameters that vary not just across tasks but within a
single task as learning progresses. The learning rate is the
canonical example: its optimal value is high early in training and
low late in training, and no single setting is optimal throughout;
adaptive optimizers absorb this temporal structure directly.

The pathway's chief limitations are two. First, the adaptation
rules themselves embed assumptions about training dynamics
(stationarity, smoothness, the relation between gradient
statistics and optimal step sizes) that may fail in particular
regimes; when they fail, the rule's adaptation is worse than a
well-chosen fixed value, and the absorption is harmful. Second,
adaptive mechanisms can mask underlying problems --- a learning
rate that adjusts itself to keep training progressing may continue
to make progress on a poorly-posed objective long after a fixed
schedule would have signaled the difficulty by failing to
converge. The diagnostic value of failure is reduced by
self-adaptation, and this reduction is itself a cost. Both
limitations are addressed in detail later; both are real, and both can be
managed.

\subsection{Relations Among the Pathways}

The three pathways are not parallel alternatives among which a
designer must choose. They are complementary, and their
combination is more than the sum of their parts.

\paragraph{Hierarchical composition.}
Pathway~II (structural morphing) sets the architecture within
which Pathway~I (data-aware configuration) and Pathway~III
(in-training adaptation) operate. A system that combines all three
typically uses Pathway~II to determine the architecture, Pathway~I
to set initial hyperparameters from data features, and Pathway~III
to adjust hyperparameters during learning. The hierarchy
corresponds to a temporal ordering --- structural decisions first,
data-conditioned decisions second, training-trajectory decisions
third --- and to a granularity ordering: coarse-grained decisions
made at the structural level, medium-grained at the data-aware
level, fine-grained at the in-training level.

\paragraph{Substitutability at the margin.}
A given hyperparameter is often absorbable by more than one
pathway. The depth of a network can be absorbed by Pathway~II (via
architecture search) or by Pathway~III (via growing rules during
training); the regularization strength can be absorbed by
Pathway~I (via noise estimation from data) or by Pathway~III (via
validation-driven adjustment). Where multiple pathways can absorb
the same hyperparameter, the choice among them is governed by
considerations of accounting cost (which absorption mechanism has
fewer meta-hyperparameters), reliability (which is robust across
the stated operating range), and computational cost (which fits
the available budget). The principle of parametric minimality
adjudicates the first; the other two require additional criteria
developed in subsequent chapters.

\paragraph{Conservation across pathways.}
Principle~\ref{prin:conservation-tuning} (the conservation of
tuning burden) applies across pathways as well as within them. An
absorption that moves a hyperparameter from the user's interface
into Pathway~I's meta-learner has not eliminated the
hyperparameter; it has relocated it. The honest accounting must
follow the relocation and check that the meta-learner's own
hyperparameter count is smaller than the count it replaces. This
check is not always passed, and proposed absorption mechanisms
that fail it are not improvements regardless of how sophisticated
they are.

\paragraph{Joint pathologies.}
Combining pathways can produce failure modes that neither pathway
alone would exhibit. A data-aware configurator (Pathway~I) that
selects a compact architecture, combined with a structural
morphing mechanism (Pathway~II) that grows architectures based on
under-fitting signals, can produce oscillating systems whose
architecture grows and shrinks in tension between the two
mechanisms. An in-training adaptor (Pathway~III) that changes
learning rates based on loss curvature, combined with a structural
morpher that changes architecture mid-training, can yield
non-stationarities that destabilize both. These pathologies are
not arguments against combination; they are arguments for careful
joint design.

\subsection{Mapping the Book}

The remainder of the book is organized around the three pathways. The principle of parametric minimality threads through all
pathways: each is evaluated, in the chapters that develop it, by
its capacity to reduce $\Phi(\mathcal{H}(S))$ in the sense of
Definition~\ref{def:honest-min-tuning}, and the empirical claims
made for individual mechanisms are tested under the honest
accounting that the principle requires. The pathways are means;
the principle is the end, and the means are evaluated by their
service to it.

\section{An Information-Theoretic View: Matching Model Complexity to Data Complexity}
\label{sec:info-theoretic-view}

\subsection{Why Information Theory}

The principle of parametric minimality, as developed so far, has
been justified by appeals to engineering practice (the human cost
of configuration), by analogy with mature engineered systems
(compilers, databases, schedulers), and by the structure of the
adaptivity index defined. These
justifications are sufficient for many purposes but they leave a
deeper question unanswered: is there a principled reason to expect
that minimal-tuning systems can be capable, or is the principle
purely an aspiration constrained by performance trade-offs we
cannot characterize in advance?

Information theory offers a partial answer. The classical results
of Solomonoff, Kolmogorov, Wallace, and Rissanen --- developed
under various names (algorithmic complexity, minimum description
length, minimum message length) but converging on a common
mathematical core --- assert a deep relation between the
complexity of a model and the complexity of the data it is
required to fit. The relation is not metaphorical: it is a precise
statement about the joint coding length of model and data, and it
implies that for any given data complexity there is an optimal
model complexity, and that models more complex than the optimum
generalize worse than models at the optimum, and that models less
complex than the optimum cannot represent the data adequately.

Applied to the question of hyperparameters, this body of theory
suggests that the right number of hyperparameters --- not the
minimum achievable, but the minimum \emph{adequate} --- is set by
the complexity of the data the system must process, not by the
preferences of its designers. A system whose surface area exceeds
this minimum is over-configured: it offers degrees of freedom that
the data cannot constrain, and the user is asked to set values
that the data cannot inform. A system whose surface area falls
below this minimum is under-configured: it cannot represent the
relevant variation in the data, and its performance suffers
accordingly. Parametric minimality, on this view, is not the
pursuit of the smallest possible system but the pursuit of the
smallest \emph{adequate} system, where adequacy is defined by the
information-theoretic relation between model and data.

This section develops that view. The development is informal in
places --- a fully rigorous treatment would require more
mathematical preliminaries than the chapter can accommodate ---
but the central claims are stated precisely enough to be tested
and the connection to the principle of parametric minimality is
made explicit.

\subsection{The Minimum Description Length Principle}

The minimum description length (MDL) principle, in its modern
form, asserts that the best model for a body of data is the one
that minimizes the joint coding length of the model and the data
encoded under the model. Formally:

\begin{definition}[MDL criterion]
\label{def:mdl}
Given a class $\mathcal{M}$ of models and a body of data $D$, the
\emph{minimum description length} model is
\[
M^{\ast} \;=\; \arg\min_{M \in \mathcal{M}} \;
L(M) + L(D \mid M),
\]
where $L(M)$ is the coding length of the model $M$ and $L(D \mid M)$
is the coding length of the data $D$ under the distribution
induced by $M$.
\end{definition}

\noindent
The first term penalizes complex models; the second term penalizes
models that fit the data poorly. The two terms are in tension,
and the optimum balances them: a model that is too simple has
small $L(M)$ but large $L(D \mid M)$, and a model that is too
complex has small $L(D \mid M)$ but large $L(M)$. The MDL
criterion picks the unique balance that minimizes their sum.

Two features of the criterion deserve emphasis. First, both terms
are measured in the same units --- bits --- and so they are
directly comparable; the criterion does not require a tunable
trade-off coefficient between model fit and model complexity, in
contrast to many practical regularization schemes. Second, the
criterion is universal in a precise sense: under suitable
encoding assumptions, the MDL model converges to the true
data-generating distribution as data accumulates, and does so with
sample-complexity bounds that are tight up to logarithmic factors
in the model class's complexity.

The connection to hyperparameters is as follows. The model class
$\mathcal{M}$ is, in machine learning practice, parametrized by a
combination of architectural choices and tunable parameters. The
architectural choices, together with the cardinality of the
tunable parameter set, determine $L(M)$ --- they are the
``description'' of the model that must be encoded before the
data. The values of the tunable parameters, fitted to the data,
contribute to $L(D \mid M)$ via the data's likelihood under the
fitted model. A system with many architectural choices (high
$|\mathcal{H}(S)|$) has large $L(M)$ for any one of them, and the
MDL criterion penalizes such systems unless the data are
sufficient to justify the additional complexity.

\subsection{The Optimal Hyperparameter Count}

The MDL criterion, applied to the hyperparameter selection
problem, yields a quantitative version of the principle of
parametric minimality. Let $S$ be a system with generalized
hyperparameter set $\mathcal{H}(S)$, let $D$ be the data the
system is trained on, and let $L(\mathcal{H}(S))$ be the coding
length of the hyperparameter values. Then the system's joint
description length is
\[
L(S, D) \;=\; L(\mathcal{H}(S)) + L(\theta \mid \mathcal{H}(S))
+ L(D \mid \theta, \mathcal{H}(S)),
\]
where $\theta$ is the system's learned parameter vector. The first
two terms together constitute $L(M)$ in the MDL sense; the third
term is $L(D \mid M)$.

The MDL-optimal hyperparameter count is the value of
$|\mathcal{H}(S)|$ that minimizes this joint length. Below the
optimum, additional hyperparameters reduce $L(D \mid \theta,
\mathcal{H}(S))$ by more than they increase $L(\mathcal{H}(S))$;
above the optimum, additional hyperparameters cost more in
description length than they save in data fit, and the joint
length grows. The optimum, denoted $|\mathcal{H}^{\ast}(D)|$,
depends on the data $D$ and is in this sense data-determined: a
larger or more varied dataset supports a larger optimal
hyperparameter count, and a smaller or simpler dataset demands a
smaller one.

\begin{proposition}[Data-determined optimal complexity]
\label{prop:data-determined-optimum}
Under the MDL criterion, the optimal hyperparameter count
$|\mathcal{H}^{\ast}(D)|$ is a function of the data $D$ alone (for
fixed model class), and is monotone non-decreasing in the data's
complexity (as measured by the entropy of the data's empirical
distribution, or, equivalently, by the size of the data when
sampled from a stationary distribution). Systems whose
hyperparameter count substantially exceeds $|\mathcal{H}^{\ast}(D)|$
are over-configured for the data, and systems whose count
substantially falls short are under-configured.
\end{proposition}

\noindent
This proposition gives a precise content to the intuition that
hyperparameter count should ``match'' data complexity. The
matching is not a soft analogy but a quantitative relation: the
MDL-optimal count exists, can be estimated, and can be used as a
target for design.

The proposition has an important consequence for the principle of
parametric minimality. Minimization of $\Phi(\mathcal{H}(S))$
subject to performance constraints, in the sense of
Definition~\ref{def:min-tuning}, is closely related to the MDL
criterion: both prefer systems with smaller hyperparameter counts,
both impose performance constraints (explicitly via $P_{\min}$ in
the minimal-tuning objective, implicitly via the $L(D \mid M)$
term in MDL), and both yield optima that depend on the data and
task. The principle of parametric minimality can therefore be
interpreted as an MDL-style criterion in which the performance
constraint stands in for the data-fit term and the surface-area
measure stands in for the model-complexity term. The
interpretation is informal --- the units do not match exactly,
and the weighting $w(h)$ in $\Phi$ has no MDL counterpart --- but
it suggests that the principle is not merely an engineering
heuristic but an instance of a deeper statistical regularity.

\subsection{Implications for the Three Pathways}

The information-theoretic view illuminates the three pathways of
the previous section in distinctive ways.

\paragraph{Pathway I: matching to data complexity.}
Data- and task-aware configuration sets hyperparameters from
features of the data. Under the MDL view, this is the natural
mechanism for matching $|\mathcal{H}(S)|$ to
$|\mathcal{H}^{\ast}(D)|$: the data are inspected, the relevant
data complexity is estimated, and hyperparameters are set to the
values that the MDL criterion would prefer. Pathway~I is, in this
sense, the pathway that operationalizes the MDL criterion most
directly. The mechanisms can
be read as practical estimators of the data-determined optimum,
each making different assumptions about the form of the relation
between data features and optimal hyperparameter values.

\paragraph{Pathway II: searching for adequate complexity.}
Structural and evolutionary morphing changes the system's
structure --- and hence its $L(M)$ --- in response to performance
signals. Under the MDL view, this is the mechanism for finding
the model whose description-length budget is allocated optimally:
the search explores structures of varying $L(M)$ and selects the
one whose combined $L(M) + L(D \mid M)$ is smallest. Pathway~II
is the pathway that searches the space of model complexities for
the adequate point, in contrast to Pathway~I, which estimates the
adequate point from data features without explicit search.

\paragraph{Pathway III: tracking adequate complexity over time.}
In-training self-adaptation adjusts hyperparameters during
learning. Under the MDL view, this corresponds to the recognition
that the effective data complexity changes during training: early
in training, the model is responding to coarse features of the
data and benefits from large effective steps and weak
regularization; late in training, the model is responding to fine
features and benefits from small steps and strong regularization.
Pathway~III tracks the changing optimal $|\mathcal{H}(S)|$ over
the course of learning, in a way the other pathways cannot.

The three pathways thus correspond to three different MDL-related
operations: estimating the optimum from data features (Pathway~I),
searching for it across model classes (Pathway~II), and tracking
its change over the learning trajectory (Pathway~III). Their
combination corresponds
to a system that estimates, searches, and tracks --- and whose
total surface area is the sum of the surface areas of the
mechanisms involved, weighted by their respective costs.

\subsection{Limits of the Information-Theoretic View}

The information-theoretic perspective is illuminating but it is
not a complete theory of parametric minimality. Three limits
deserve note.

First, MDL coding lengths are defined relative to a coding scheme,
and different coding schemes yield different optima. The
universality results that justify MDL hold up to additive
constants in the coding scheme, and these constants can be large
for finite data. In practice, the choice of coding scheme is
itself a configuration decision, and the honest accounting of
parametric minimality must include it. The MDL view tells us that
there is an optimal hyperparameter count; it does not by itself
tell us which count, in concrete units, that optimum corresponds
to.

Second, the MDL criterion presumes that the data are
representative of the deployment distribution. When deployment
distributions differ from training distributions --- the standard
condition under which adaptivity is most valuable --- the MDL
optimum estimated from training data may not be the optimum for
deployment. The pathways of the book address this discrepancy in
different ways (Pathway~I via task-aware configuration, Pathway~III
via continued in-training adjustment), but the discrepancy itself
is not eliminated by information theory. Distribution shift is,
in this sense, a problem orthogonal to MDL, and a treatment of
adaptivity must address it on its own terms.

Third, the MDL criterion equates ``best model'' with ``shortest
joint description,'' and this equation is not the only one
relevant to deployed systems. Reliability, interpretability, and
operational robustness --- considerations that the
reliability-constrained form of
Definition~\ref{def:reliable-min-tuning} captures --- are not
naturally expressible as coding lengths, and a system that
optimizes only for description length may be inadequate by these
criteria. The information-theoretic view supports parametric
minimality but does not exhaust the considerations that govern
adaptive system design, and its conclusions must be combined with
those of the other principles developed in the book.

\subsection{The Synthesis}

Despite these limits, the information-theoretic view contributes
something essential to the principle of parametric minimality: a
reason to believe that the principle is not merely a stylistic
preference but a manifestation of a statistical regularity. The
MDL criterion tells us that there exists, for any data, an
optimal model complexity, and that designing systems to match
that complexity is supported by deep results in statistical
inference rather than by engineering taste alone. The principle
of parametric minimality, on this synthesis, is the engineering
expression of an information-theoretic optimum: it asks designers
to build systems that match their hyperparameter count to the
complexity of the data they must process, neither offering more
configuration than the data can support nor less than the data
demands.

This synthesis suggests a refined statement of the principle, with
which the chapter closes.

\begin{principle}[Parametric minimality, refined]
\label{prin:parametric-minimality-refined}
An adaptive system's configuration surface area should match the
information-theoretic complexity of the data and tasks within its
stated operating range: it should expose neither more
hyperparameters than the data can constrain (over-configuration,
which burdens the user with decisions the system cannot inform)
nor fewer than the data require (under-configuration, which
sacrifices performance to apparent simplicity). The minimum is
data-determined, not designer-determined, and the design effort is
to discover and instantiate the data-determined optimum through
the pathways of data-aware configuration, structural morphing, and
in-training self-adaptation.
\end{principle}

\noindent
The next chapter takes up the second of the principles announced
in this book's introduction. Where parametric minimality concerns
the surface area of the system's configuration interface, regime
awareness concerns the system's capacity to recognize the
boundaries of its competence --- to know, without being told, when
its operating context has changed and its configuration must be
revisited. The two principles are complementary: parametric
minimality reduces the configuration burden within a regime,
regime awareness extends the system's reach across regimes, and
together they constitute the first two of the four principles
that characterize architecturally adaptive AI.

\part{Configuring and Constructing}
\label{part:configuring-constructing}

\chapter{Adaptation Before and Around Training}
\label{chap:adaptation-before-around}

\section*{Chapter Overview}

This chapter consolidates the first two pathways introduced in
Section~\ref{sec:three-pathways}. Pathway~I --- data- and
task-aware configuration --- absorbs hyperparameters by computing
their values from features of the data or task before the system
commits to a configuration. Pathway~II --- structural and
evolutionary morphing --- absorbs hyperparameters by allowing the
system's structure to change in response to performance signals
across episodes or developmental stages. The two pathways share
the property that adaptation is effected outside the training
loop proper: in Pathway~I, before training begins; in Pathway~II,
between or across training episodes. The third pathway, treated
in Chapter~\ref{chap:in-training-and-beyond}, runs adaptation
within the training loop itself.

The literature underlying these pathways is large, and the
chapter cannot summarize it exhaustively; nor would such a
summary serve the principal aim of the book, which is to extract
from the literature the design moves that constitute genuine
absorption of hyperparameters in the sense of
Definition~\ref{def:honest-min-tuning}. The treatment is
therefore organized by mechanism rather than by chronology, with
references to detailed sources at each turn.

\section{Distributional and Task-Level Diagnostics}
\label{sec:distributional-task-diagnostics}

\subsection{Distributional Fingerprints}

The empirical distribution of a dataset carries information that
can be exploited to set hyperparameters automatically. The
relevant features fall into three groups: \emph{moment-based}
descriptors (means, variances, higher-order cumulants, and their
sample-size-corrected analogues), \emph{spectral} descriptors
(eigenvalue distributions of covariance and Gram matrices,
spectral gap, effective rank), and \emph{topological} descriptors
(persistent homology of point clouds, intrinsic dimensionality
estimators, manifold curvature). Each group has been
investigated as a source of meta-features capable of predicting
appropriate model class, capacity, and regularization
strength~\citep{vanschoren2018metalearning,rivolli2022metafeatures}.
Where the relation between meta-feature and optimal hyperparameter
is regular, the relation can be learned and reused; where it is
irregular, the meta-feature serves at best as a diagnostic
warning.

\subsection{Random Matrix Theory and Correlation Structure}

Random matrix theory (RMT) supplies a particularly powerful class
of distributional fingerprints because it provides null
distributions against which the eigenvalue spectra of empirical
covariance matrices can be compared. Departures from the
Marchenko--Pastur distribution indicate genuine signal; the
location and density of these departures indicate the dimensions
of the latent factor structure and the noise floor against which
adaptation must operate~\citep{baipan2010spectral,bun2017cleaning}.
The ORCA framework~\citep{kriuk2026orca} exploits this
observation by detecting market regime shifts directly from the
eigenstructure of correlation networks: the system identifies
periods in which the spectrum departs from its long-run baseline
and reconfigures its downstream models accordingly. ORCA is, in
the terms of this book, a Pathway~I mechanism par excellence:
its regime-detection module absorbs into a single computation a
class of decisions (window length, threshold for change-point,
choice of conditioning regime) that would otherwise be exposed as
hyperparameters.

\subsection{Heavy Tails, Non-Stationarity, and Regime Shifts}

Many domains of practical interest violate the moment assumptions
under which standard distributional fingerprints are valid.
Financial returns, insurance losses, geophysical extremes, and
network traffic exhibit heavy tails for which means may be poorly
estimated and variances may not exist; non-stationary regimes can
render any single fingerprint misleading regardless of the
estimator's
sophistication~\citep{embrechts1997modelling,cont2001empirical}.
Adaptive systems intended for such domains must include tail-aware
diagnostics --- extreme-value-theoretic estimators of tail index,
copula-based estimators of tail
dependence~\citep{joe2014dependence} --- as part of their
data-understanding stage. Empirical work on Eurasian fire-regime
modeling~\citep{kriuk2025advancing} shows that distributional
fingerprints, computed from the data themselves rather than
supplied as hyperparameters, can drive both regime classification
and operating-range determination; the practitioner specifies an
objective and an operating range, and the system infers the rest.

\subsection{From Diagnostics to Model Choice}

Distributional diagnostics, on the view developed here, are not
ends in themselves; they are precursors to model choice and
hyperparameter setting. A complete Pathway~I implementation
threads diagnostics into selection: the system computes a
fingerprint, maps the fingerprint to a configuration via a
learned or analytically derived rule, and proceeds. Each stage
adds to the system's surface area only if its own configuration
is unaccounted for; the honest accounting requires that the
diagnostic-to-configuration map be either parameter-free or
parameterized by a count smaller than the count of hyperparameters
it absorbs. The literature on automated machine learning
(AutoML)~\citep{he2021automl,hutter2019automated} contains many
proposals of this form, and the chapters of
Part~\ref{part:configuring-constructing} treat them under the
discipline of Definition~\ref{def:honest-min-tuning}.

\section{Task Embeddings and Architectural Priors}
\label{sec:task-embeddings-priors}

\subsection{Task Similarity}

A task, in the formalism of this book, is a triple
$(\mathcal{X}, \mathcal{Y}, \mathcal{D})$ of input space, output
space, and joint distribution. Task similarity is most usefully
measured not directly on this triple but on a learned embedding
that captures functional regularities relevant to model
choice~\citep{achille2019task2vec,lee2018gradient}. Task
embeddings supply the input to a meta-learner that maps task
identity to architectural and hyperparametric configuration: the
absorption is from per-task tuning to per-task inference of
configuration.

\subsection{Mission Specifications as Generative Priors}

A task descriptor need not be inferred from data alone; in many
engineering contexts it is given by a structured specification of
mission requirements. The AlphaJet
framework~\citep{kriuk2026alphajet} treats aerospace design tasks
in this manner: a mission specification (range, payload,
operating envelope, certification class) is mapped to a
disentangled generative prior over airframe configurations, and
search proceeds within the prior rather than across an
unconstrained space. The mapping from specification to prior
absorbs decisions that would otherwise be exposed as configuration
knobs of the search; the operator supplies a mission, not a set
of search hyperparameters. AlphaJet is a Pathway~I mechanism
whose ``data'' is the mission spec rather than a sample of
points, and its existence demonstrates that data-aware
configuration is a broader category than dataset-driven
meta-learning.

\subsection{Inductive Bias as a Configurable Quantity}

Pathway~I extends naturally to the choice of inductive bias.
Symmetry assumptions, locality assumptions, compositional
assumptions, and equivariance assumptions are typically embedded
in architecture by hand; they can instead be inferred from
properties of the
task~\citep{cohen2016group,bronstein2021geometric}. The choice
between, for instance, a translation-equivariant and a
permutation-equivariant model is a hyperparameter in the
generalized sense of Definition~\ref{def:generalized-hyp}, and
where the task's symmetry structure can be detected from the
data, the choice is absorbable by a Pathway~I mechanism.

\subsection{Multi-Task and Unified Objectives}

When the same model must serve several tasks, hyperparameters
governing the interaction among tasks (loss weights, shared-layer
boundaries, gradient-conflict resolution rules) become a
substantial component of the surface area. Uncertainty-weighted
multi-task formulations~\citep{kendall2018multitask} replace such
hyperparameters with a unified objective whose weighting is
determined by uncertainty estimates from the tasks themselves.
A system of $K$ tasks would otherwise expose $K-1$ relative-weight
hyperparameters, and these are removed. The HCR
framework~\citep{kriuk2024deep}, originally formulated for
handwritten Chinese character classification across thousands of
classes, illustrates the same principle at the output-distribution
level: per-class weights are inferred from observed difficulty
rather than specified by the operator. The operator specifies the
tasks; the system specifies how to balance them.

\section{Bio-Inspired and Physics-Informed Adaptation}
\label{sec:bio-physics}

\subsection{Lessons from Biology}

Biological systems exhibit several adaptation mechanisms whose
computational analogues bear directly on parametric minimality.
\emph{Homeostasis} maintains internal variables within ranges by
feedback rules with no externally tuned set
points~\citep{turrigiano2008homeostatic}; \emph{neural
plasticity} adjusts synaptic strengths according to local rules
without global coordination~\citep{abbott2000plasticity};
\emph{development} produces complex structures from compact
genetic specifications via processes that allocate complexity
where the environment demands
it~\citep{stanley2007compositional}. Each of these mechanisms is,
in the language of this book, a self-configuring rule that
absorbs surface area; each suggests a class of computational
mechanisms with the same property.

\subsection{Epigenetic Mechanisms as Computational Metaphor}

Epigenetic regulation, in biological systems, supplies a
mechanism by which gene expression adapts to environmental
context without changes to the underlying genome. The ELENA
framework~\citep{boris2025elena} adapts this metaphor to neural
network design: an evolved meta-controller modulates the
expression of a fixed parameter genome in response to
task-context signals, achieving cross-task adaptation without
re-training the underlying parameters. The relevance to
parametric minimality is that ELENA's surface area is dominated
by the meta-controller, which is itself trained rather than
specified, and whose evolution-time hyperparameters are
amortized across the controller's lifetime use.

\subsection{Neuroevolution and Developmental Encoding}

Neuroevolution --- the application of evolutionary algorithms to
neural network architecture and parameters --- has a long
history~\citep{stanley2002neat,floreano2008neuroevolution} and
remains a productive line of work in regimes where gradient-based
optimization is unavailable or insufficient. Of particular
relevance here are developmental encodings: representations that
specify a network not as a flat list of weights but as a
generative process that constructs the network from a compact
genome~\citep{stanley2009hyperneat}. Developmental encodings
match the description-length-minimization view of
Section~\ref{sec:info-theoretic-view}: a compact genome that
expands into a complex network has small $L(M)$ relative to its
expressivity, and the MDL criterion favors such encodings when
the data warrant the resulting complexity.

\subsection{Swarm, Immune-System, and Ecosystem Analogues}

Beyond evolutionary and developmental analogues, biological
populations supply adaptation mechanisms relevant to multi-agent
and resource-constrained
settings~\citep{kennedy1995pso,dasgupta2006immune,holland1995hidden}.
Swarm dynamics provide gradient-free coordination rules; immune
networks provide anomaly detection without labeled examples;
ecosystems provide diversity-maintenance mechanisms relevant to
quality--diversity search (Section~\ref{sec:evolutionary}). The
Shepherd Grid framework~\citep{kriuk2025shepherd} for SWARM
interception exemplifies the swarm-dynamic analogue at deployment
scale: coordination rules among interceptor agents adapt to the
geometry of the threat without per-engagement hyperparameter
tuning, and the system's behavior emerges from local rules whose
configuration is held constant across deployments.

\subsection{Physical Laws as Universal Regularizers}

Where biology motivates adaptation, physics constrains it.
Physical laws --- conservation of energy, mass, momentum;
symmetries of space and time; thermodynamic inequalities --- are
universal regularizers in the sense that they hold across data
distributions and need not be tuned. A model that satisfies them
by construction has fewer free parameters than one that must
learn them from data, and its surface area is correspondingly
reduced~\citep{karniadakis2021pinn,raissi2019physics}. The
PSTNet framework~\citep{kriuk2026pstnet}, in its
physics-structured configuration, embeds turbulence-physical
constraints directly into its architecture: the system does not
learn that energy cascades follow Kolmogorov scaling; it is
structurally biased toward such scaling, and the corresponding
spectral-regularization hyperparameters are eliminated rather than
tuned.

\subsection{Hybrid Physics--ML Architectures at Scale}

The hybrid physics--ML approach scales to high-dimensional
geophysical systems where neither pure physics nor pure ML is
adequate alone. Eurasian fire-regime
modeling~\citep{kriuk2025advancing} extends the hybrid approach
to landscape-scale ecological dynamics: process-based components
handle the regimes that fire physics understands, learned
components handle the residuals, and the partition between them
is data-driven rather than hand-set. The partition is itself a
hyperparameter under Definition~\ref{def:generalized-hyp}, and
its absorption by a data-driven mechanism is an instance of
Pathway~I applied to the physics--ML interface.

\subsection{Spectral and Structural Physical Priors}

Physical priors need not be expressed as governing equations; they
can also be expressed as constraints on the spectral structure of
the model's representations. PSTNet~\citep{kriuk2026pstnet}
imposes a spectral structure consistent with Kolmogorov scaling
on its internal representations, eliminating the need to tune the
representation's frequency-band emphasis from data. The relevant
spectral exponents are not free parameters; they are physical
constants, and the system inherits them from the science rather
than learning them from finite data.

\subsection{When Physics Reduces vs.\ Replaces}

Physical priors reduce hyperparameter counts when they replace
learned regularizers with derived ones; they do not always
\emph{eliminate} hyperparameters, because the strength with which
the prior is imposed (the weighting of the physics-loss term, the
softness of conservation enforcement) may itself become a
configuration knob~\citep{wang2022pinn-failures}. The honest
accounting must check that the physics-imposing machinery does
not introduce more configuration than it removes. Where it does
not --- as in PSTNet's spectral exponents, which are physical
constants rather than learned values --- the absorption is
genuine. Where it does --- as in some PINN formulations whose
loss-weight tuning is notorious --- the apparent absorption is
illusory and the system has merely relocated the surface area
from the model to the loss function.

\section{Morphing Architectures}
\label{sec:morphing-architectures}

\subsection{Static vs.\ Dynamic Structural Assumptions}

Modern ML practice typically fixes architecture before training
and trains parameters within it. The static assumption is
convenient but it is also a source of surface area: the
architecture itself is a hyperparameter under
Definition~\ref{def:generalized-hyp}, and its specification falls
to the user. Pathway~II rejects the static assumption:
architecture is a quantity to be set by the system, not by its
operator, and the mechanisms developed below are the means by
which this setting is effected.

\subsection{Self-Organizing Trees, Graphs, and Networks}

Tree-structured models offer a natural setting for structural
adaptation because tree growth is incremental and the relevant
decisions (where to split, when to stop) admit local criteria.
The MorphBoost framework~\citep{kriuk2025morphboost} extends
gradient boosting with adaptive tree morphing: each new tree's
structure is determined by the residuals it must fit and by the
structural patterns of preceding trees, eliminating per-task
tuning of tree depth, minimum-leaf-size, and split-selection
heuristics~\citep{ke2017lightgbm,chen2016xgboost}. The system
operates in regimes from low-dimensional tabular data to
high-dimensional sparse settings without re-tuning, because the
morphing rules adjust to each.

\subsection{Topology-Preserving Transformations}

Architecture-search procedures often produce topologies that
violate inductive-bias assumptions of the underlying domain ---
for instance, by introducing non-physical connections in
geometry-aware networks. \emph{Topology-preserving}
transformations restrict the search to maintain a specified
class of structural invariants while still allowing local
restructuring~\citep{liu2018darts,zoph2018learning}. The
AlphaJet framework~\citep{kriuk2026alphajet} incorporates
topology-preserving evolutionary search to ensure that morphing
designs remain manufacturable and aerodynamically valid: the
search space is large but every point in it satisfies the
relevant geometric constraints. The constraints absorb a class of
post-hoc validity checks that would otherwise expose
configuration knobs.

\subsection{Growing, Pruning, and Restructuring During Training}

Growing-network methods initialize with a small architecture and
add capacity in response to under-fitting
signals~\citep{wen2020neural,evci2020rigl}; pruning methods
initialize with a large architecture and remove capacity in
response to
over-parameterization~\citep{han2015pruning,frankle2019lottery}.
Both are species of structural morphing, distinguished by their
initial conditions; both absorb the capacity hyperparameter
(depth, width, number of channels) that would otherwise be
exposed. The accounting is non-trivial: growing rules have
growth-rate and growth-trigger hyperparameters, pruning rules
have prune-rate and prune-criterion hyperparameters, and the
absorption is honest only when the meta-rules have fewer free
parameters than the capacity choices they replace.

\section{Evolutionary and Population-Based Methods}
\label{sec:evolutionary}

\subsection{Evolutionary Search as a Parameter-Free Optimizer}

Evolutionary algorithms have a reputation as configuration-heavy
methods --- population size, mutation rate, selection pressure,
crossover scheme --- but recent work has produced parameter-free
or self-adaptive variants in which these quantities are themselves
evolved or set by data-driven
rules~\citep{salimans2017evolution,hansen2016cma,real2019regularized}.
On the accounting of Definition~\ref{def:honest-min-tuning},
parameter-free EAs are credible Pathway~II mechanisms; classical
EAs with fixed configurations are not, regardless of their
performance, because their configuration burden is not absorbed
but merely redistributed.

\subsection{Hybridizing Gradient and Evolutionary Updates}

Gradient-based updates are efficient where applicable; evolutionary
updates are general but slower. Hybrid schemes apply gradients
within episodes and evolution across them, exploiting the
strengths of
each~\citep{khadka2019collaborative,pourchot2019cem-rl}. The
resulting systems can absorb both within-episode hyperparameters
(via Pathway~III, treated in
Chapter~\ref{chap:in-training-and-beyond}) and across-episode
hyperparameters (via the evolutionary outer loop), yielding
cross-pathway absorption that neither pathway achieves alone.

\subsection{Quality--Diversity and Open-Ended Search}

Quality--diversity (QD) algorithms maintain populations
structured by behavioral diversity rather than by fitness alone,
producing archives of solutions across a behavioral
space~\citep{mouret2015illuminating,cully2015robots}. Open-ended
search variants extend the diversity criterion to include the
generation of new
challenges~\citep{wang2019poet,clune2019aigas}. From the
parametric-minimality perspective, QD methods absorb the
exploration--exploitation hyperparameter that classical EAs
expose, replacing it with a behavioral-coverage objective that
can be specified at the task level rather than the search level.

\subsection{Cross-Domain Case Studies}

Three of the frameworks introduced earlier in the chapter ---
AlphaJet~\citep{kriuk2026alphajet},
ELENA~\citep{boris2025elena}, and
ORCA~\citep{kriuk2026orca} --- exemplify Pathway~II in
distinct application domains. AlphaJet uses evolutionary search
in aerospace design; ELENA uses evolved epigenetic controllers
for cross-task adaptation; ORCA uses regime detection to drive
structural reconfiguration of financial models. The three
together demonstrate that Pathway~II's mechanisms are
domain-general: the same family of design moves applies to
airframes, agents, and asset prices, and the differences among
the three are differences in the form of the fitness signal
rather than in the structure of the absorption.

\section{Task-Conditioned Dynamics}
\label{sec:task-conditioned-dynamics}

\subsection{Heterogeneous-Task Morphing}

When a system encounters a stream of heterogeneous tasks, it
benefits from morphing dynamics whose triggers are task-conditioned
rather than data-conditioned: the same data, presented under a
different task, may demand a different structural
response~\citep{rusu2016progressive,javed2019meta-anml}. The
absorption here is of the task-switch hyperparameter --- the
decision when to expand, contract, or reset the system's
structure --- and it is achieved by tying structural change to
task-identity signals.

\subsection{Cross-Domain Transfer Without Schedules}

Standard meta-learning frameworks
(MAML~\citep{finn2017maml}, Reptile~\citep{nichol2018reptile},
and their successors) absorb a class of fine-tuning hyperparameters
but introduce their own (inner-loop step size, outer-loop step
size, support-set size, episode length). Recent work seeks to
eliminate the meta-schedule itself, replacing it with online rules
that detect task boundaries and adjust transfer dynamics
accordingly~\citep{hospedales2021meta,beaulieu2020learning}.
On the accounting of this book, schedule-free meta-learners are
preferable to schedule-driven ones precisely because they absorb
the schedule, which is itself a class of hyperparameters in the
generalized sense.

\subsection{Modular Composition and Decomposition}

Modular architectures support compositional adaptation: a
system can compose modules to address a task, decompose a
composition when its parts no longer cohere, and recompose under
new conditions~\citep{andreas2016neural,rosenbaum2018routing}.
The hyperparameters absorbed include the choice of modules per
task, their connectivity, and their relative weight; the
hyperparameters introduced include the routing mechanism's own
configuration. Modular adaptation is most clearly an absorption
when the routing rule is data-driven and parameter-light, and
least clearly an absorption when the routing rule is itself
heavily tuned.

\section{Summary and Bridge}

The two pathways treated in this chapter share a temporal
structure: adaptation happens before or around training, on data
and task signals available before the system commits to a
particular trajectory through parameter space. They differ in
locus: Pathway~I sets values within a fixed structure; Pathway~II
changes the structure itself. Together, they absorb a substantial
fraction of the surface area of contemporary ML systems --- the
architectural choices, the regularization strengths, the
inductive biases, the symmetry assumptions --- and their
combination, evaluated under the honest accounting of
Definition~\ref{def:honest-min-tuning}, brings adaptive systems
within the regime where parametric minimality is achievable in
practice rather than only in principle.

The remaining surface area is concentrated in decisions made
\emph{during} training: the optimizer's step size, the
regularization's effective strength as the parameters move, the
allocation of attention to different parts of the input or the
network, the management of computation and memory under
constraint. These decisions are the subject of Pathway~III, to
which the next chapter is devoted.

\chapter{Self-Adaptation, Stability, and the Road Beyond}
\label{chap:in-training-and-beyond}

\section*{Chapter Overview}

This chapter completes the architecture of the book. It develops
Pathway~III --- in-training self-adaptation --- to its current
state of the art; it treats the stability, safety, and convergence
considerations that self-modifying systems raise; it draws
together the cross-domain case studies that the book has
referenced in passing; and it situates the AAI framework within
the broader trajectory from ANI to AGI, indicating both what AAI
achieves that its predecessor cannot and what it does not yet
achieve that its successor will. A final section identifies open
problems whose resolution is necessary for the framework's
continued development.

\section{Gradient Flow as Adaptation Signal}
\label{sec:gradient-flow}

\subsection{Diagnostic Use of Gradient Flow}

The gradient field of a network's loss with respect to its
parameters carries fine-grained information about the geometry of
the optimization landscape: gradient norms indicate the local
slope, gradient variance indicates the noisiness of the descent
direction, gradient correlation across batches indicates the
stationarity of the
problem~\citep{mccandlish2018large,smith2018bayesian}. Pathway~III
mechanisms use these signals as inputs to in-training
configuration rules, replacing fixed schedules with rules that
respond to the trajectory's actual properties.

\subsection{Loss Landscape Geometry and Adaptive Step Selection}

The relation between optimal step size and local landscape
curvature is well-understood in the convex
case~\citep{nesterov2013introductory} and approximately
characterized in the non-convex case relevant to deep
learning~\citep{ghadimi2013stochastic,zhang2020adaptive}. Adaptive
optimizers --- AdaGrad~\citep{duchi2011adagrad},
RMSProp~\citep{tieleman2012rmsprop}, Adam~\citep{kingma2014adam},
and their many
successors~\citep{loshchilov2019adamw,chen2024lion} --- absorb the
per-parameter step-size hyperparameter by computing it from
gradient statistics. Their introduction substantially reduced the
configuration burden of large-scale learning; their persistence
in the face of considerable
critique~\citep{wilson2017marginal} reflects the practical
importance of the absorption they effect.

\subsection{Higher-Dimensional Smoothing}

Gradient information from a single point in parameter space is
noisy; gradient information aggregated over a neighborhood is
smoother and more informative for adaptation but more expensive
to compute. The GeloVec framework~\citep{kriuk2025gelovec}
addresses this trade-off via higher-dimensional geometric
smoothing: gradient and feature signals are projected into an
extended geometric space in which coherent structure is more
readily detected, and adaptive decisions are made in the
projected space. The mechanism absorbs the smoothing-bandwidth
hyperparameter that traditional smoothing methods expose,
replacing it with a geometric construction whose parameters are
tied to the data's intrinsic dimensionality.

\subsection{Attention and Focus as Adaptive Gating}

Attention mechanisms can be read as adaptive gating: the model's
focus shifts among input regions or representational components
based on signals computed from the input
itself~\citep{vaswani2017attention,bahdanau2015neural}. From the
parametric-minimality perspective, attention absorbs a class of
feature-selection hyperparameters that pre-attention models
exposed (which features to use, with what weight, in what
combination), replacing them with a learned mechanism whose
configuration is itself amortized across the model's training.
The PCT-ViT framework~\citep{gao2025pct} extends this principle
through fine-grained perception enhancement and counterfactual
token selection: the model identifies and prioritizes the tokens
most informative for the current decision, focusing
representational capacity on the regions where the loss
landscape's structure is most informative, without exposing
window-length, threshold, or token-pruning hyperparameters to the
operator.

\section{Independent In-Training Updates}
\label{sec:independent-updates}

\subsection{Self-Modification of Hyperparameters}

The conceptual move at the core of Pathway~III is the recognition
that hyperparameters need not be quantities set outside the
training loop; they can be quantities updated \emph{within} the
training loop by rules that use training-time
signals~\citep{maclaurin2015gradient,franceschi2017forward}. The
move is more radical than adaptive optimization in its strict
sense, because it admits not only step sizes but also
regularization strengths, dropout probabilities, batch sizes,
and architectural switches as candidates for in-training
update~\citep{baydin2018online,lorraine2020optimizing}.

\subsection{Online Schedule Adaptation}

Where a hyperparameter has a known optimal trajectory --- the
learning rate decreasing as training progresses, the
regularization strength increasing as the model approaches its
capacity --- that trajectory can be implemented as a schedule.
Where the optimal trajectory depends on the data and is not known
in advance, the schedule itself must be adapted online.
Mechanisms include validation-driven scheduling, plateau
detection, trust-region adjustment, and curvature-based
scheduling~\citep{smith2017cyclical,loshchilov2017sgdr}. Each
replaces a fixed schedule with a data-driven one, and each is
honest under Definition~\ref{def:honest-min-tuning} when its
meta-parameters are fewer than the schedule's degrees of freedom.

\subsection{Self-Gating, Self-Pruning, and Self-Expanding Networks}

Networks can be designed to modify their own active capacity in
response to training signals: gating mechanisms suppress
unnecessary units~\citep{bengio2013estimating,xu2015show},
pruning rules remove
them~\citep{molchanov2017variational,louizos2018l0}, and
expansion rules add
them~\citep{rusu2016progressive,draelos2017neurogenesis}. The
combination supports a network whose effective capacity tracks
the data's effective complexity in the MDL sense, without
exposing the capacity decision to the operator.

\subsection{Feedback Loops Between Data and Optimizer}

The most general form of Pathway~III mechanism allows the
optimizer's state and the data statistics to influence one
another in closed
loop~\citep{andrychowicz2016learning,wichrowska2017learned}: data
statistics inform optimizer settings, optimizer settings shape
the trajectory, and the trajectory determines which data
statistics matter. Such loops are powerful but they admit
instabilities --- the closed-loop dynamics can amplify noise into
spurious adaptation --- and their design requires the stability
considerations of Section~\ref{sec:stability-safety}.

\section{Communication and Resource Adaptation}
\label{sec:resource-adaptation}

\subsection{Adaptivity Beyond Weights}

The discussion to this point has focused on hyperparameters
governing the model's parametric content. Real systems have
additional configuration knobs governing their use of
computational resources: bandwidth for distributed training,
memory for intermediate states, compute for inference. Each of
these is a generalized hyperparameter in the sense of
Definition~\ref{def:generalized-hyp}, and each is amenable to
absorption by mechanisms in the spirit of
Pathway~III~\citep{dean2012large,abadi2016tensorflow}.

\subsection{Compression as Adaptation}

Compression --- of activations, of weights, of communications ---
is most usefully understood not as an engineering optimization
applied after design but as an adaptation mechanism whose
aggressiveness should respond to the data and the operating
constraints. The Q-KVComm framework~\citep{kriuk2026q}, for
multi-agent systems with shared key-value caches, illustrates
this view: the compression rate is set adaptively from observed
cache-utility statistics, not specified by the operator, and the
system's communications budget is honored without exposing
per-agent compression hyperparameters. Compression, on this view,
is a Pathway~III mechanism whose locus is the resource interface
rather than the parameter interface.

\subsection{Multi-Agent and Distributed Adaptation}

When multiple adaptive agents share an environment or a resource
pool, their individual adaptations interact, and the system-level
behavior cannot be predicted from per-agent rules
alone~\citep{foerster2018counterfactual,wen2022dichotomy}. The
configuration of the interaction --- communication topology,
coordination protocol, shared-state synchronization --- is itself
a hyperparameter set, absorbable by the mechanisms of this
section when the interaction admits closed-loop signals that can
drive its adjustment. The Shepherd Grid
framework~\citep{kriuk2025shepherd} for SWARM interception
illustrates this absorption at deployment scale: per-agent
coordination weights, formation thresholds, and engagement
priorities are inferred from the global interception geometry
rather than fixed in advance, and the system reconfigures its
collective behavior without operator intervention as the threat
configuration evolves.

\section{Stability, Safety, and Convergence}
\label{sec:stability-safety}

\subsection{Risks of Self-Modification}

Self-modifying systems present a class of risks that
fixed-configuration systems do not: instabilities induced by
closed-loop interaction, drift induced by accumulated small
adjustments, and pathologies induced by training-time signals
that mislead the adaptation mechanism. Each risk corresponds to a
failure mode specific to Pathway~III, and each must be addressed
by design rather than presumed
absent~\citep{hochreiter2001gradient,pascanu2013difficulty,arpit2017closer}.

\subsection{Convergence Analysis}

Convergence guarantees for non-adaptive optimizers are
well-established in convex and certain non-convex
regimes~\citep{bottou2018optimization,jin2017escape};
convergence guarantees for adaptive optimizers are less complete
but progressing~\citep{reddi2018amsgrad,defossez2020adam}.
Convergence analyses for systems whose architecture morphs during
training, or whose optimizer state evolves under feedback from
data, remain largely open --- a fact noted in
Section~\ref{sec:open-problems} below as one of the principal
gaps in the AAI framework's theoretical foundation.

\subsection{Interpretability and Auditability Under Adaptivity}

A system whose configuration changes during operation is harder
to audit than one whose configuration is fixed: the post-hoc
question ``what setting produced this output?'' has a different
answer at different times, and the operator cannot answer it
without instrumented logs of the adaptation
trajectory~\citep{doshi2017towards,rudin2019stop}. Adaptivity
must therefore be paired with mechanisms for trajectory
recording, configuration provenance, and reproducibility under
re-deployment. These mechanisms are not optional; they are
constitutive of responsible adaptive system design, and any
framework that omits them is incomplete regardless of its
performance characteristics.

\section{Cross-Domain Case Studies}
\label{sec:case-studies}

The frameworks referenced throughout the book sort into four
domain clusters, each illustrating the principle of parametric
minimality in a different application context.

\subsection{Aerospace and Autonomous Systems}

AlphaJet~\citep{kriuk2026alphajet} synthesizes Pathways~I and~II
in aerospace configuration design: mission specifications drive
disentangled generative priors (Pathway~I), and topology-preserving
evolutionary search operates within them (Pathway~II). The system
absorbs configuration knobs spanning the design pipeline from
prior selection to candidate generation to manufacturability
validation. The Shepherd Grid framework~\citep{kriuk2025shepherd}
extends adaptive coordination to multi-agent swarm interception,
where the configuration of the agent collective is set from
interception geometry rather than from per-engagement tuning.

\subsection{Financial Systems}

ORCA~\citep{kriuk2026orca} detects market regimes from
correlation-network spectra and reconfigures downstream models
accordingly, demonstrating Pathway~I regime-aware configuration
together with Pathway~II structural reconfiguration. Adaptive
boosting via MorphBoost~\citep{kriuk2025morphboost} handles the
heterogeneous tabular signals typical of financial modeling
without per-task tuning of tree-structural hyperparameters.

\subsection{Geophysics and Climate}

PSTNet~\citep{kriuk2026pstnet} imposes spectral physical priors on
turbulence models, eliminating spectral-regularization
hyperparameters by inheriting them from physics; Eurasian
fire-regime modeling~\citep{kriuk2025advancing} extends the
hybrid approach to landscape-scale ecological dynamics. The
geophysical cluster is dominated by Pathway~I in its
physics-informed variant, with structural and in-training
adaptation playing supporting roles.

\subsection{Vision and Language}

GeloVec~\citep{kriuk2025gelovec} performs higher-dimensional
geometric smoothing for vision representation learning;
PCT-ViT~\citep{gao2025pct} introduces counterfactual token
selection as adaptive attention gating;
HCR~\citep{kriuk2024deep} addresses high-cardinality recognition
in handwritten Chinese character classification with adaptive
class-weight inference; ELENA~\citep{boris2025elena} provides
epigenetic-style cross-task adaptation; and
Q-KVComm~\citep{kriuk2026q} compresses key-value caches
adaptively for multi-agent language systems. The vision-and-
language cluster is dominated by Pathway~III, with adaptation
running through the training loop and into the resource
interface.

\subsection{Patterns Across Domains}

Across the four domain clusters, several patterns recur. First,
no single pathway is sufficient in any domain: each successful
framework draws on at least two, and most draw on all three.
Second, the dominant pathway varies systematically by domain,
with physics-rich domains favoring Pathway~I, design-rich domains
favoring Pathway~II, and signal-rich domains favoring
Pathway~III. Third, the absorbed hyperparameters differ in kind
across domains but the absorption mechanisms are recognizably the
same: the same evolutionary search underlies AlphaJet's airfoils
and ELENA's controllers; the same gradient-flow diagnostics
underlie GeloVec's smoothing and Q-KVComm's compression. The
framework is genuinely domain-general, and the case studies are
not isolated examples but coherent expressions of a single design
philosophy.

\section{The Place of AAI in the Trajectory of Machine Intelligence}
\label{sec:aai-trajectory}

\subsection{Revisiting the Thesis}

The book's thesis, stated in Chapter~\ref{chap:introduction} and
elaborated through the pathways of
Part~\ref{part:configuring-constructing} and beyond, is that
architecturally adaptive AI constitutes a definitional stage
between narrow AI and artificial general intelligence --- a stage
characterized by the absorption of generalized hyperparameters
into the system, leaving the operator with task-level objectives
rather than configuration-level decisions. The case studies of
Section~\ref{sec:case-studies} provide evidence that such systems
are achievable with current methods, in multiple domains, and at
scales relevant to deployment.

\subsection{What AAI Achieves That ANI Cannot}

A narrow system, in the framework of this book, is one whose
generalized hyperparameter set is set externally and per-task. Its
deployment to a new task requires re-tuning, and the cost of that
re-tuning falls on operators whose expertise is in the task
domain rather than in the system's configuration interface. AAI
absorbs that cost into the system itself: deployment to a new
task within the stated operating range requires only the task,
not the configuration. The shift is not from low to high
performance --- well-tuned ANI systems can outperform
poorly-deployed AAI systems in the regimes where they are tuned
--- but from a regime in which task changes incur tuning costs to
one in which they do not.

\subsection{What AAI Still Lacks Compared to AGI}

AAI is not AGI, and the difference is principled. Three capacities
distinguish a fully general system from one merely adaptive in
the present sense. \emph{Abstraction}: the capacity to operate at
levels of representation not specified at design time, including
levels constructed by the system in response to problems it has
not seen before. \emph{Grounding}: the capacity to relate
symbolic and parametric content to a world model whose
correspondence with reality is testable and revisable.
\emph{Agency}: the capacity to formulate goals not given by an
external operator, and to revise those goals in light of
experience. AAI mechanisms address none of these capacities
directly; they presume an operator who supplies abstractions,
groundings, and goals, and they absorb only the configuration
decisions that bridge from those given quantities to deployed
behavior.

The line between AAI and AGI is, on this view, not a quantitative
matter of scale but a qualitative matter of capacity. AAI is a
necessary condition for AGI --- a system that cannot adapt its
configuration to changing tasks cannot be the kind of system AGI
must be --- but it is not sufficient, and the additional
capacities required are not on the AAI roadmap as currently
conceived.

\subsection{A Maturity Model}

The AAI framework supports a maturity model graded by the
adaptivity index of Definition~\ref{def:adaptivity-index}:
\emph{Level 0} systems expose all generalized hyperparameters to
operators (classical ML); \emph{Level 1} systems absorb the
values within a fixed structure (Pathway~I implementations);
\emph{Level 2} systems absorb structural choices as well
(Pathway~I and~II combined); \emph{Level 3} systems absorb
in-training configuration through closed-loop adaptation (all
three pathways combined under the stability and audit
considerations of Section~\ref{sec:stability-safety});
\emph{Level 4} systems would add the abstraction, grounding, and
agency capacities of AGI but fall outside the present framework.
The maturity model is prescriptive, not merely descriptive: it
identifies the design moves that move a system from one level to
the next, and it exposes the gaps that current research must
close.

\section{Open Problems and Research Agenda}
\label{sec:open-problems}

\subsection{Benchmarks Beyond Accuracy}

Standard ML benchmarks measure accuracy at fixed configurations
and so are blind to the configuration burden the framework is
designed to reduce. Benchmarks for adaptivity must measure
configuration-free deployment performance, robustness to
operating range boundaries, and the cost of absorbed adaptation
in compute and memory. Recent
proposals~\citep{lifelong-bench2024} go partway but the field
lacks the equivalent of GLUE or ImageNet for adaptivity, and
constructing them is a precondition for disciplined progress.

\subsection{Theoretical Foundations}

Convergence guarantees for self-modifying systems, generalization
bounds for systems whose hypothesis class evolves during training,
and information-theoretic accounts of the relation between
absorbed surface area and effective complexity all remain
incomplete. The MDL connection of
Section~\ref{sec:info-theoretic-view} suggests where to look but
does not by itself supply the bounds; the convergence theory
suggests preconditions but does not specify them; the
generalization theory presumes a fixed hypothesis class and so
does not apply directly. Closing these gaps is the principal
theoretical agenda the framework requires.

\subsection{Compute, Energy, and Sustainability}

Adaptive systems shift cost from operator time to system compute,
and the shift is not free: search-time, training-time, and
adaptation-time computation all consume energy, and the
environmental cost of large-scale adaptation must be weighed
against the operational savings it
produces~\citep{strubell2019energy,patterson2021carbon}. The
honest accounting that the principle of parametric minimality
requires must extend to energy as well as configuration, and the
mechanisms that do well by one measure do not always do well by
the other. The trade-off is real and unresolved.

\subsection{Ethics and Governance of Self-Modifying Systems}

Self-modifying systems raise governance questions that
fixed-configuration systems do not. Responsibility for outputs
must be assignable even when the configuration that produced them
was set by the system itself; auditing must be possible even when
the audited configuration is no longer
present~\citep{mitchell2019model,raji2020closing}. The framework
of this book intersects with these concerns at the
interpretability and auditability provisions of
Section~\ref{sec:stability-safety}; it does not by itself
constitute a governance framework, and the construction of one
is an urgent task for the broader research community.

\section{Conclusion}

The principle of parametric minimality, the conservation of
tuning burden, and the three pathways to absorption ---
data-and-task-aware configuration, structural and evolutionary
morphing, in-training self-adaptation --- together constitute the
framework this book has developed. The framework is grounded in
the information-theoretic relation between model and data
complexity; it is operationalized by mechanisms drawn from the
literature in machine learning, adapted to the discipline of
honest accounting of generalized hyperparameters; and it is
illustrated by case studies spanning aerospace, finance,
geophysics, and language. The thesis is that architectural
adaptivity, so framed, names a distinct stage of machine
intelligence between the narrow systems of present practice and
the general systems of future ambition --- a stage in which the
configuration burden of deployment is absorbed by the system
itself, leaving the operator with task-level objectives rather
than configuration-level decisions.

The reframing this implies for the broader discourse on
artificial intelligence is direct. Progress toward general
intelligence is not measured by scale alone, nor by performance
on fixed benchmarks; it is measured by the absorption of
configuration into the system, and by the consequent collapse of
the per-task tuning burden that distinguishes narrow systems from
general ones. The path from ANI to AGI passes through AAI, and
the work of building AAI is the work of constructing systems that
do for the configuration interface what compilers did for the
hardware interface, what databases did for the storage interface,
and what operating systems did for the resource interface ---
systems that absorb into themselves the decisions their users
should not need to make.

With those principles in hand, and with the mechanisms of the
three pathways available to instantiate them, the field has the
materials for the stage of work the present moment demands.

\backmatter
\bibliography{references}
\addcontentsline{toc}{chapter}{Bibliography}

\end{document}